\documentclass[10pt,twocolumn,letterpaper]{article}

\usepackage[pagenumbers]{cvpr} 

\definecolor{cvprblue}{rgb}{0.21,0.49,0.74}
\usepackage[pagebackref,breaklinks,colorlinks,allcolors=cvprblue]{hyperref}
\usepackage{multirow}

\title{NovelGS: Consistent Novel-view Denoising via Large \\
Gaussian Reconstruction Model}

\author{
 Jinpeng Liu$^{1}$, Jiale Xu\thanks{Project Lead.} $^{,2}$, Weihao Cheng$^{2}$, Yiming Gao$^{2}$, Xintao Wang$^{2}$,  Ying Shan$^{2}$, Yansong Tang\thanks{Corresponding author.} $^{,1}$ \\
 $^{1}$Tsinghua Shenzhen International Graduate School, Tsinghua University \\
 $^{2}$ARC Lab, Tencent PCG
 }
 
\begin{document}
 
\newcommand{\mrka}[1]{{\colorbox{red!30}{#1}}}
\newcommand{\mrkb}[1]{{\colorbox{red!20}{#1}}}
\newcommand{\mrkc}[1]{{\colorbox{red!10}{#1}}}

\maketitle

\begin{abstract}
We introduce NovelGS, a diffusion model for Gaussian Splatting (GS) given sparse-view images. Recent works leverage feed-forward networks to generate pixel-aligned Gaussians, which could be fast rendered. Unfortunately, the method was unable to produce satisfactory results for areas not covered by the input images due to the formulation of these methods.  In contrast, we leverage the novel view denoising through a transformer-based network to generate 3D Gaussians. Specifically, by incorporating both conditional views and noisy target views, the network predicts pixel-aligned Gaussians for each view. During training, the rendered target and some additional views of the Gaussians are supervised. During inference, the target views are iteratively rendered and denoised from pure noise. Our approach demonstrates state-of-the-art performance in addressing the multi-view image reconstruction challenge. Due to generative modeling of unseen regions, NovelGS effectively reconstructs 3D objects with consistent and sharp textures. Experimental results on publicly available datasets indicate that NovelGS substantially surpasses existing image-to-3D frameworks, both qualitatively and quantitatively. We also demonstrate the potential of NovelGS in generative tasks, such as text-to-3D and image-to-3D, by integrating it with existing multiview diffusion models. We will make the code publicly accessible. 
\end{abstract}    
\section{Introduction}
The automation of 3D content creation holds substantial promise across various domains such as digital gaming, virtual reality, and cinematic production. Core methodologies, including image-to-3D and text-to-3D, offer considerable advantages by substantially reducing the dependency on manual labor by professional 3D artists. Some work~\citep{dreamfusion, magic3d, dream3d, dreamgaussian, prolificdreamer, fantasia3d, dreamcraft3d,latent-nerf} generate 3D assets by iteratively distilling image generative models. However, methods based on Score Distillation Sampling (SDS) necessitate prolonged optimization periods per asset, often extending to several hours. Due to the limited understanding of 3D concepts in 2D diffusion models, maintaining 3D consistency is challenging. As a result, these methods are prone to producing geometric artifacts, such as the multi-faced Janus~\citep{dmv3d} and issues related to content drift.

\begin{figure}[t]
    \centering
    \includegraphics[width=0.47\textwidth]{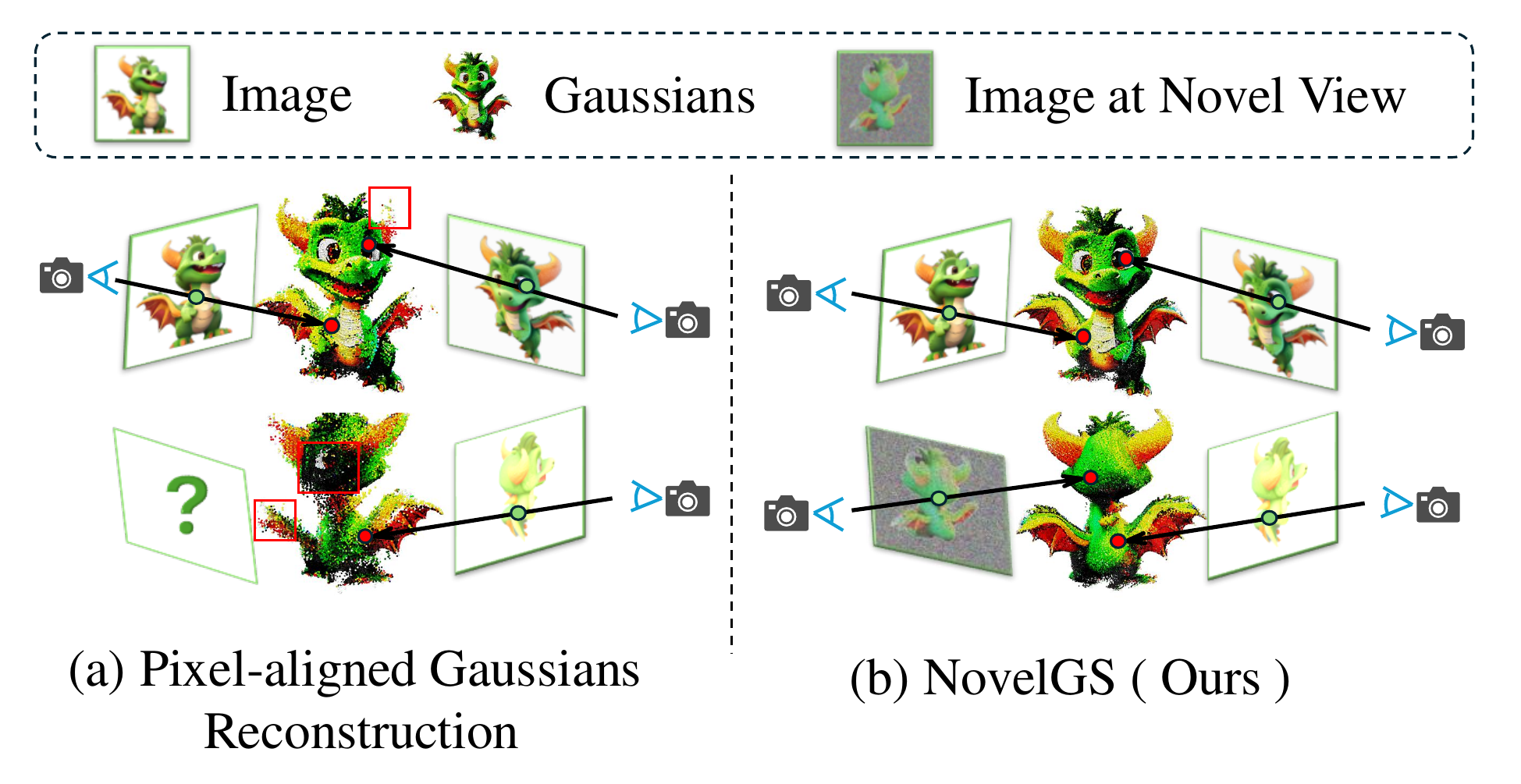}
    \vspace{-8pt}
    \caption{\textbf{ Comparison of pixel-aligned Gaussians reconstruction models and NovelGS.} (a) Most existing models~\citep{lgm, grm, gs-lrm} translate the input pixels into pixel-aligned Gaussians~\citep{grm} based on camera rays. (b) Conversely, we propose to denoise novel view images via the large Gaussian reconstruction model where the unseen parts of the objects could be reconstructed consistently. }
    \label{fig:intro}
    \vspace{-15pt}
\end{figure}

\begin{figure*}[htbp]
    \centering
    \includegraphics[width=1.0\linewidth]{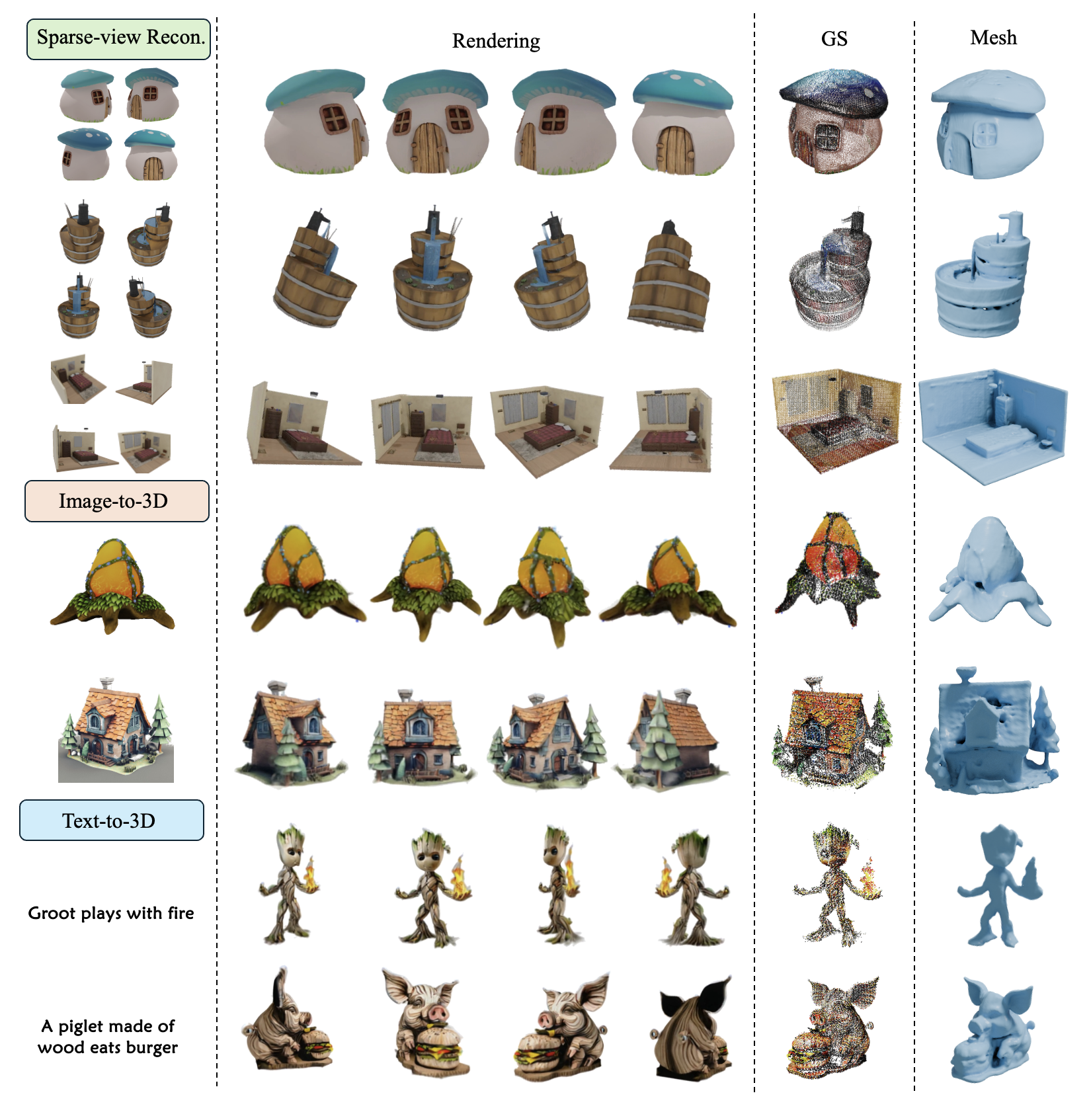}
    \vspace{-20pt}
    \caption{\textbf{High-fidelity 3D assets} produced by~\textbf{NovelGS}. It's designed for sparse-view reconstruction and operates in conjunction with various complementary tools, including text-to-image generation~\cite{stable-diffusion}, and image-to-multiview modeling~\cite{zero123plus}. This collaborative framework facilitates the generation of text-to-3D (bottom) and image-to-3D (center), as well as the reconstruction of real-world objects (top).}
    \vspace{-15pt}
    \label{fig:enter-label}
\end{figure*}

With the advent of large 3D datasets~\citep{objaverse, objaverse-xl} and implicit 3D representations~\citep{nerf, eg3d}, some studies~\citep{lrm, gs-lrm, CRM, instant3d} propose utilizing transformer-based models to map images into triplane features in a feed-forward manner. They then render novel views using volume rendering techniques~\citep{nerf}. While these methods are flexible, they result in dense computations during rendering, which can be time-consuming. For instance, rendering a 2-second (60 frames) video takes approximately 1.5 minutes on a single NVIDIA A100 GPU. To enhance user-friendliness, some studies~\citep{instantmesh, meshlrm} propose combining similar frameworks with the Marching Cubes algorithm~\cite{flexible-cube, marching-cube} to generate 3D meshes directly. However, this approach is challenging and unstable during training, and the rendering quality is suboptimal.

3D Gaussians~\cite{gs-splatting} features fast rendering speeds with explicit representation. As shown in ~\Cref{fig:intro} (a), some studies~\cite{lgm, grm, gs-lrm, geolrm, gaussiancube} utilize stacks of transformer or U-Net models to map images to pixel-aligned Gaussians. However, they tend to poorly generalize to novel views that are not covered by input views. Because they correspond the pixel points of the image to spatial locations based on the camera's perspective, the results tend to be poor and inconsistent for areas not illuminated by the camera.

In this paper, we propose NovelGS, a 3D Gaussian diffusion model conditioned on a few input images. NovelGS utilizes a transformer-based denoising network, which is fed with not only condition views but also a number of noisy views as shown in ~\Cref{fig:intro} (b). These target views are preset for unseen regions, to generate parts not covered by condition views. The network then predicts pixel-aligned 3D Gaussians for all these views. During training, we expect that clean and noisy views are rendered from the predicted Gaussians and supervise them with $L2$ and $LPIPS$ loss. During inference, we initialize target views with pure noise and step-by-step denoise them by the network, and we obtain the final Gaussians from the last denoising step. 
Specifically, we introduce the denoise of the novel view in the reconstruction process to ensure the consistent visual effect of the invisible part (see ~\cref{exp:ab}). At the same time, our model structure is flexible and can accept various combinations of different numbers and positions of noisy views and clean views befitting the application scenarios. The model is conditioned on the diffusion time step, allowing it to manage varying noise levels throughout the diffusion process.

We trained NovelGS on multi-view images of Objaverse~\cite{objaverse} and evaluated the performance on the Google Scanned Objects~\cite{gso} and OmniObject3D~\cite{omni3d}. By integrating novel-view denoising, our model not only outperforms existing methods with the same input views but also makes it possible to handle unbalanced input images, which couldn't cover enough parts of the objects. When paired with text-to-image~\cite{sd} and image-to-multi-view image models~\cite{zero123plus}, NovelGS achieves outperforming quality for text and single image-to-3D object generation. Experimental results demonstrate the state-of-the-art performance of our method in sparse-view reconstruction benchmarks.

\section{Related Work}
\subsection{Reconstruction Models}
Reconstructing 3D from multi-view images is a long-standing problem in computer vision. Traditional methods rely on fitting, which usually requires a dense set of images, such as NeRF~\cite{nerf} and Gaussian Splatting~\cite{gs-splatting}. Learning-based methods use neural networks to predict 3d representations from sparse images, e.g., MVSNeRF, PixelNeRF, NerFormer, SRT, MCC~\cite{pf-lrm, CRM, meshlrm, instantmesh, instant3d, triposr}. Among these methods, large reconstruction models (LRMs)~\cite{lrm} demonstrate strong generalization ability on open-world images. By training on large-scale datasets~\cite{objaverse, objaverse-xl}, LRMs effectively maps a single image to triplanes~\cite{eg3d} via a transformer-based network. Instant3D extends LRMs to a text-to-3d method. It first uses diffusion model to generate multi-view images from text, and then uses LRM to predict triplanes from the images. As an implicit representation, Triplanes are not only effective for novel-view synthesis but also can be extracted into high-quality mesh~\cite{meshlrm, instantmesh}. Some work~\cite{lgm, grm, gs-lrm} explores Gaussians~\cite{gs-splatting} as the 3D representation. LGM~\cite{lgm} and GRM~\cite{grm} utilizes an asymmetric U-Net and a transformer network to predict and fuse 3D Gaussians, respectively. GeoLRM~\cite{geolrm} proposes to utilize occupancy grid prediction to predict geometry-aware objects. GS-LRM~\cite{gs-lrm} validates the feasibility of the paradigm in a large-scale scene dataset. Compared with these methods, our NovelGS utilizes the transformer-based novel-view diffusion model to denoise noisy novel-view images utilizing conditional information from known images, explicitly exploring unseen parts of the 3D object. 

\subsection{3D Generation}
The field of generative models has experienced significant advancements, particularly with the development of Generative Adversarial Networks (GANs)~\cite{gan} and Diffusion Models~\cite{diffusion, wonder3d, 3dgen}, which have demonstrated substantial efficacy in image and video generation~\cite{taming-gan, stable-diffusion, make-a-video}.  In the context of 3D generation, 3D GANs are utilized to generate 3D-aware asserts~\cite{pi-gan, 3d-aware-gan, epigraf, giraffe, hologan} in early time, while they are hard to train, leading to limited performance. Although some works utilize 3D diffusion models~\cite{pointe, shape, auto3d, 3dgen, 3d-nerf} to replace 3D GANs with direct 3D supervision for 3D assert generation, the quality and diversity of their results are significantly lower compared to the performance of DMs in 2D space. This discrepancy is partly due to the computational challenges of scaling diffusion network models from 2D to 3D and the limited availability of 3D training data~\cite{shapenet} previously. DMV3D~\cite{dmv3d} utilizes multi-view diffusion to denoise images, while it's hard to extend to the scene and NeRF~\cite{nerf} is time-consuming for rendering. Some studies~\cite{meshgpt} utilize an autoregressive model~\cite{gpt-1} to generate meshes directly. While mesh representation is challenging to encode and not GPU-friendly, this leads to instability during the training stage and suboptimal rendering quality. In contrast, NovelGS employs an efficient Gaussian representation and novel view denoising, resulting in improved efficiency and stability for both training and inference.
\section{Method}
\begin{figure*}[t]
    \centering
    \includegraphics[width=1.0\textwidth]{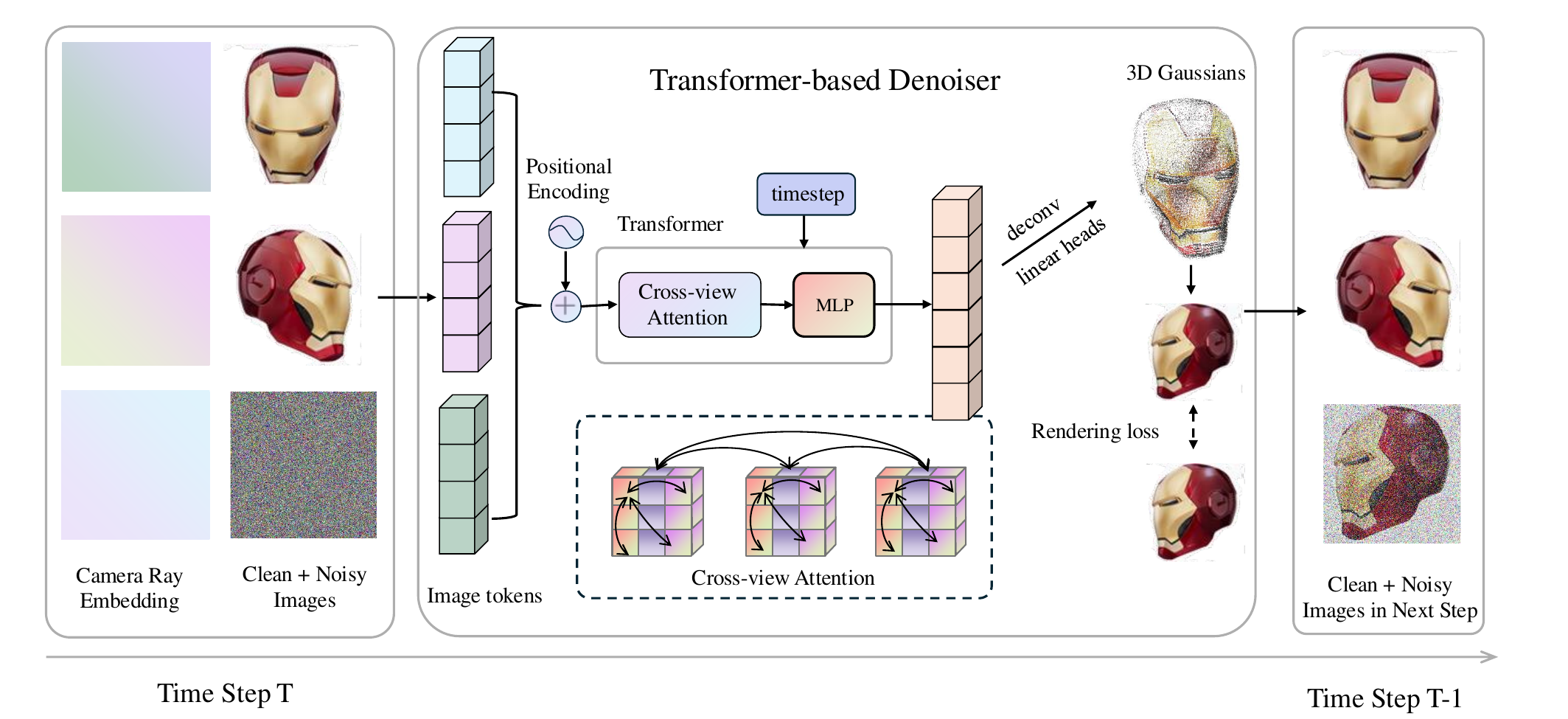}
    \caption{\textbf{Pipeline of NovelGS model.} We utilize a large transformer-based network to denoise noisy view images for 3D reconstruction. \textbf{During inference}, we initialize target views with pure noise. Then we concatenate the camera ray embedding (Plücker rays) and images (two clean views and one noisy view in the figure to reduce clutterness; four clean views and one noisy view in main experiments) as the input. Then we utilize the denoiser to predict the Gaussians and render the image from the noisy view. After that, we add noise to the noisy view images to timestep T-1. We loop this process until we get the final 3D Gaussians. \textbf{During training}, we add noise to the noisy view images based on the timestep and utilize the denoiser to predict 3D Gaussians. We train the denoiser module by rendering loss.}
    \label{fig:pipeine}
    \vspace{-10pt}
\end{figure*}

In this section, we introduce our NovelGS model, which is designed to reconstruct high-quality 3D assets from sparse-view images. Our approach leverages a diffusion framework that effectively denoises images from noisy views through 3D Gaussian reconstruction and rendering, facilitating consistent 3D generation (see ~\Cref{pip:denoise}). Additionally, we propose a transformer-based denoiser for generating 3D Gaussians~\cite{lgm, grm, gs-lrm}, which conditions on both the timestep and clean images. This enables precise and controllable 3D reconstruction (see ~\Cref{pip:condition}). The final output of the denoising process is a set of 3D Gaussians, culminating in the generated 3D model. The loss functions employed in our model are detailed in ~\Cref{pip:loss}.

\subsection{Model Architecture} \label{pip:denoise}
The pipeline of our model is shown in the~\Cref{fig:pipeine}. During the training phase, our method utilizes a set of images $\{I^i\}^{m+n}_{i=1}$ along with their corresponding camera ray embeddings $\{R^i\}^{m+n}_{i=1}$ as input, where $m$ and $n$ represent the number of clean and noisy images, respectively. We add different levels of noise to noisy view images $\{I^i\}^{m+n}_{i=m+1}$ based on the timestep $T$. Moreover, a transformer-based denoiser predicts 3D Gaussians $G$. Finally, we render several images from the 3D Gaussians and supervise the model by rendering loss. In the inference stage, we initialize the noisy view image $\{I^i\}^{m+n}_{i=m+1}$ with pure noise and concatenate it with clean view images $\{I^i\}^{m}_{i=1}$. Then we concatenate the set of images with their camera ray embeddings as the input of denoiser. Moreover, the denoiser outputs 3D Gaussians, and we render the Gaussians in noisy views. After that, we add noise to the noisy view images to timestep $T-1$ and replace the noisy view images at timestep $T$. Finally, they will serve as the input for the next diffusion sampling step until we get the final 3D Gaussians at timestep 0.

\textbf{Gaussians and Camera Embedding.} Gaussian splatting~\cite{gs-splatting} represents 3D scene with a set of 3D gaussians, which are efficient for rendering. Specifically, each Gaussian is defined by a center $\mathbf{x} \in R^3$, a scaling factor $\mathbf{s} \in R^3$, and a rotation quaternion $\mathbf{q} \in R^4$. Additionally, an opacity value $\mathbf{\alpha} \in R$ and spherical harmonics (SH) coefficients $\mathbf{c} \in R^D$, with $D$ denoting the number of SH bases, are maintained for rendering. These parameters can be collectively denoted by $\Theta$, with $\Theta = \{x_i, s_i, q_i, {\alpha}_i, c_i\}$ representing the parameters for the i-th Gaussian. Following previous methods~\cite{pixelsplat, viewset, grm, lgm}, we use the Plücker ray embedding to encode the camera poses to get camera embedding:
\vspace{-5pt}
\begin{equation}
    f_i = \{o_i \times d_i, d_i \} \label{eq:ray}
\end{equation}
where $d_i$ is the ray direction, and $o_i$ is the ray origin. Each pixel of the output feature map is treated as a 3D Gaussian inspired by splatter image~\cite{splatter}. Consequently, for each input view the model predicts a Gaussian attribute map $H \in R^{H \times W \times C}$ of $C$ channels, corresponding to depth, rotation, scaling, opacity, and the DC term of the SH coefficients. Then $m+n$ views of Gaussian attribute are contacted together, generating a total of $(m+n) * H * W$ 3D Gaussians. Finally, we could render different images from any viewpoint with these 3D Gaussians $G$.

\textbf{Input Posed Image Tokenization.} 
NovelGS employs a streamlined tokenizer for posed images, drawing inspiration from the Vision Transformer~\cite{meshlrm} and MeshLRM~\cite{meshlrm}. Specifically, we concatenate the camera ray embedding with the RGB pixel values, resulting in a 9-channel feature map. This feature map is then divided into non-overlapping patches, which are linearly transformed to serve as input for the transformer. Although the Plücker coordinates inherently encode spatial information, we add additional positional embeddings following VIT which is different from MeshLRM. Because we want our model to be more sensitive to the position of the novel view. It is noteworthy that our image tokenizer is considerably simpler than those used in previous large reconstruction models (LRMs), which often rely on a pre-trained DINO ViT~\cite{dino} for image encoding. Because DINO is primarily optimized for intra-view semantic reasoning, whereas 3D reconstruction predominantly requires inter-view low-level correspondences~\cite{meshlrm}.

\textbf{Transformer-based Denoiser.} 
We concatenate multi-view image tokens with learnable triplane (positional) embeddings and input them into a sequence of transformer blocks~\cite{transformer}. Each block is composed of cross-view self-attention and multilayer perceptron (MLP) layers, with layer normalization applied before both layers and residual connections are incorporated. This deep transformer network facilitates extensive information exchange among all tokens, effectively modeling intra-view and inter-view relationships. The noisy image tokens, now contextualized by all condition views, are subsequently decoded into clean 3D tokens. Then we utilize transposed convolution to upsample the features. From the upsampled features $F$, we predict the Gaussian attribute maps for pixel-aligned Gaussians using separate linear heads. These attribute maps are subsequently unprojected along the viewing ray based on the predicted depth. This process allows for the rendering of a final image $I^i$, and an alpha mask $M^i$ (used for supervision) at an arbitrary camera view through Gaussian splatting.

\subsection{Time Step and Image Condition} \label{pip:condition}
\textbf{Time Step Condition.} Inspired by DiT~\cite{dit, instant3d}, we employ the \textit{adaLN-Zero} module to incorporate the timestep condition. In each cross-view attention module, the timestep is injected to handle inputs with varying noise levels.

\textbf{Image Condition.} 
To enhance the adaptability of our model, we adopt an approach where the initial $m$ views $\{I^1, I^2, ..., I^m\}$ in the denoiser input are kept free of noise to serve as conditioning images. Meanwhile, diffusion and denoising processes are applied to the remaining $n$ views. This strategy enables the denoiser to effectively reconstruct missing pixels in the noisy, unseen views by leveraging information from the input views, analogous to the image inpainting task, which has been demonstrated to be feasible with 2D denoising models~\cite{stable-diffusion}. Moreover, to improve the generalizability of our image-conditioned model, we generate 3D Gaussians within a coordinate frame aligned with the conditioning views and render additional images using poses relative to these conditioning views. Specifically, we normalize all camera positions together so that the position of the first condition image view resides at $(0, y, 0)$.

\subsection{Loss Function} \label{pip:loss}
During the training stage, we render images from random T supervision views using the predicted 3D Gaussians and minimize the image reconstruction loss and mask loss. Furthermore, we utilize perceptual image patch similarity loss~\cite{lpips} to make the training stage more stable. $\{I_i|i=1,2,..., H\} $ represent the ground-truth views, and $\{{\hat{I}}_i|i=1,2,..., H\}$ represent the predict views rendered by the predict Gaussian splats. $\{M_i|i=1,2,..., H\} $ represent the ground-truth mask, and $\{{\hat{M}}_i|i=1,2,..., H\}$ represent the predicted mask rendered by the predicted Gaussian splats. So our loss function is :
\begin{equation}
    \mathcal{L} = \frac{1}{T}\sum_{i=1}^{T}(\mathcal{L}_{img}(I_i, \hat{I}_i ) + \mathcal{L}_{mask}(M_i, \hat{M}_i )),
\end{equation}
\begin{equation}
    \mathcal{L}_{img}(I_i, \hat{I}_i ) = ||I_i - \hat{I}_i ||_2 + \lambda \cdot \mathcal{L}_{LPIPS}(I_i, \hat{I}_i ),
\end{equation}
\begin{equation}
    \mathcal{L}_{mask}(M_i, \hat{M}_i ) = ||M_i - \hat{M}_i ||_2,
\end{equation}

where $\mathcal{L}_{LPIPS}$ represent the perceptual image patch similarity loss, $\lambda$ is the weight of it . Note that $H$ is larger than $(m+n)$ because our model could supervise more views than input views for better performance.

\section{Experiments}
\subsection{Implementation Details}
\textbf{Training Data.} Our training dataset is composed of multi-view images rendered from the Objaverse~\cite{objaverse} dataset. For each object in the dataset, we render 512 × 512 images from 32 random viewpoints. To ensure high-quality training data, we applied a thorough filtering process to curate a subset of objects that meet specific criteria (See Supplementary). By applying these filtering criteria, we curated a high-quality subset consisting of approximately 270,000 instances from the initial pool of 800,000 objects in the Objaverse dataset. This rigorous selection process ensures that our model is trained on data that is both diverse and representative of high-quality 3D objects, thereby enhancing the robustness and accuracy of the generated 3D reconstructions.

\textbf{Evaluation Data.} We utilize two public datasets following InstantMesh~\cite{instantmesh}: Google Scanned Objects (GSO)~\cite{gso} and OmniObject3D (Omni3D)~\cite{omni3d}. To evaluate the visual quality of the generated 3D asserts, we created the image evaluation sets for both GSO and Omni3D datasets. For the GSO dataset, which comprises approximately 1,000 objects, we randomly selected 300 objects to constitute the evaluation set. For the Omni3D dataset, we chose 28 common categories and then selected the first 5 objects from each category (totaling 130 objects, as some categories contain fewer than 5 objects) as the evaluation set. For each object, we rendered 21 images along an orbiting trajectory with uniform azimuths and varying elevations of \{30°, 0°, -30°\}. This systematic evaluation approach allows us to assess the visual fidelity and quality of the 3D Gaussians generated by NovelGS. By leveraging multiple views and varying angles, we ensure a comprehensive evaluation that captures the nuanced details of the reconstructed objects.

\begin{figure*}[htbp]
    \centering
    \includegraphics[width=1.0\linewidth]{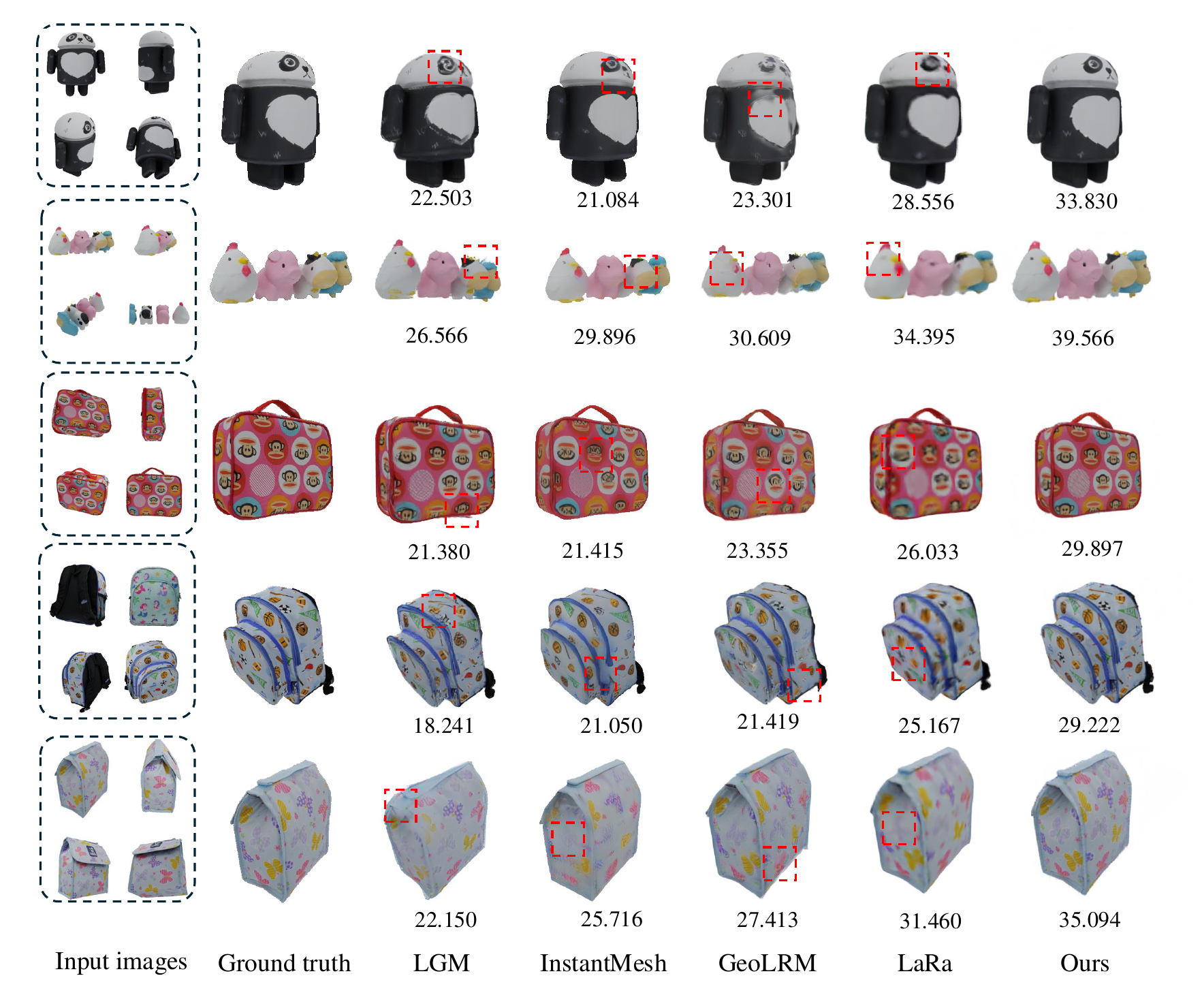}
    \vspace{-30pt}
    \caption{Visual comparisons to previous methods. The four-view input images are displayed in the leftmost column, while novel view renderings are compared on the right. Previous methods struggle to reconstruct high-frequency details and thin structures consistently. In contrast, our NovelGS demonstrates significantly improved performance in these scenarios. The PSNRs are provided beneath each image.}
    \label{fig:main}
\end{figure*}

\textbf{Training Settings.} The training process is composed of two stages. In the first stage, we pre-train the model with a resolution of 256 $\times$ 256 and a batch size of 6 in each GPU for several epochs. We utilize the AdamW optimizer~\cite{adamw} with an initial learning rate of 4e-4, which is decayed via cosine annealing after 3000 steps. In the second stage, we finetune the model with a resolution of 512 $\times$ 512 and a batch size of 2 in each GPU. We use the same optimizer~\cite{adamw} as the first stage with an initial learning rate of 4e-5. At each training step of both stages, we sample a set of 8 images (from 32 renderings) as a data point, from which we randomly select 4 clean views, 1 noisy view, and 3 supervision views independently. To optimize GPU memory usage, deferred back-propagation~\cite{deferred} and memory-efficient attention~\cite{xFormers} are employed. The model is trained on 16 NVIDIA A100 GPUs with gradient accumulation set to 8. It requires approximately two weeks to complete the training stages.

\subsection{Results and comparisons}

\textbf{Quantitative results.} In the main experiments, we select 4 clean view images and 1 noisy view image as default. We report the quantitative results of sparse view reconstruction on different evaluation sets as shown in ~\Cref{table:recon_gso} and ~\Cref{table:recon_omni}, respectively. For each metric, we highlight the top two results among all methods, and a deeper color indicates a better result. The quantitative evaluation of 2D novel view synthesis metrics indicates that NovelGS significantly outperforms the baseline models in terms of Structural Similarity Index (SSIM)~\cite{ssim} and Peak Signal-to-Noise Ratio (PSNR) ~\cite{lpips}. This superior performance suggests that NovelGS generates outputs with enhanced quality. Notably, the Learned Perceptual Image Patch Similarity (LPIPS) of NovelGS is marginally lower than that of the top-performing baseline. This observation implies that the perception of novel views generated by NovelGS exhibits slight deviations from the ground truth in human views. Because it will predict a novel view based on known input images, attributed to the ``dreaming" process inherent in the novel view diffusion process. Our model tries to image the unknown parts of the object that are more conscious of the true structure of the object. At the same time, it maintains consistency across multiple viewpoints rather than ignoring details to make the image look sensible in human views compared to the InstantMesh, as shown in the fourth row at ~\Cref{fig:main}. We believe prioritizing the consistently detailed structure of objects is imperative in the reconstruction tasks.

\begin{table}[htbp]
\vspace{-5pt}
\centering
\caption{Evaluation results on the GSO dataset. The best and the second-best scores are marked as \mrkb{red} and \mrkc{light red}. $\uparrow$ represents the higher the better, and $\downarrow$ represents the lower the better.}
\vspace{-5pt}
\begin{tabular}{l|ccc}
\hline
    & \multicolumn{3}{c}{Google Scanned Objects~\cite{gso}}  \\
    & PSNR $\uparrow$   & SSIM $\uparrow$ & LPIPS $\downarrow$  \\ \hline
LGM ~\cite{lgm} & 24.923        & 0.907        & 0.093   \\
InstantMesh~\cite{instantmesh} & 25.124        & 0.924       & \mrkb{0.059}   \\
GeoLRM~\cite{geolrm} & 25.389        &  0.918      &  0.083    \\ 
LaRa~\cite{lara} & \mrkc{28.910}        & \mrkc{0.940}        & 0.091  \\ \hline
Ours    &  \mrkb{31.303}      &   \mrkb{0.946}      &  \mrkc{0.065}  \\ \hline
\end{tabular}
\label{table:recon_gso}
\end{table}
\begin{table}[htbp]
\vspace{-15pt}
\centering
\caption{Evaluation results on Omni3D dataset.}
\vspace{-5pt}
\begin{tabular}{l|ccc}
\hline
& \multicolumn{3}{c}{OmniObject3D~\cite{omni3d}} \\
& PSNR $\uparrow$   & SSIM $\uparrow$ & LPIPS $\downarrow$ \\ \hline
LGM ~\cite{lgm}  & 24.229     & 0.913    & 0.081    \\
InstantMesh~\cite{instantmesh}   & 24.292     & 0.929    & \mrkb{0.053}    \\
GeoLRM~\cite{geolrm} & 24.289     & 0.922   &  0.083   \\ 
LaRa~\cite{lara}    &  \mrkc{28.434}    & \mrkc{0.943}    &  0.084   \\ \hline
Ours    &  \mrkb{31.195}    &  \mrkb{0.945}  & \mrkc{0.067}  \\ \hline
\end{tabular}
\label{table:recon_omni}
\end{table}

\textbf{Quantitative results.} As illustrated in~\Cref{fig:main}, to compare our NovelGS with other baselines qualitatively, we select several objects from the GSO evaluation set and obtain the sparse-view recon results. For each generated, we visualize the images of the rendering from the same viewpoints. NovelGS consistently produces visually consistent appearances, whereas baseline methods often manifest distortions in the synthesized novel views. Specifically, the NeRF-based method (InstantMesh) prefers a smooth texture, which leads to blurring on some details, as shown in the third and fifth rows of the ~\Cref{fig:main}. While other feed-forward pixel-aligned Gaussian reconstruction models would ignore some uncovered or slightly covered parts by the input view as shown in the~\Cref{fig:main}.

\subsection{Ablation}\label{exp:ab}
The key design in our method is the utilization of noisy views. We analyze our approach regarding the necessity of the noisy views, the number of noisy views, and the different positions of the noisy view and clean views.

\begin{figure}
    \centering
    \includegraphics[width=1.0\linewidth]{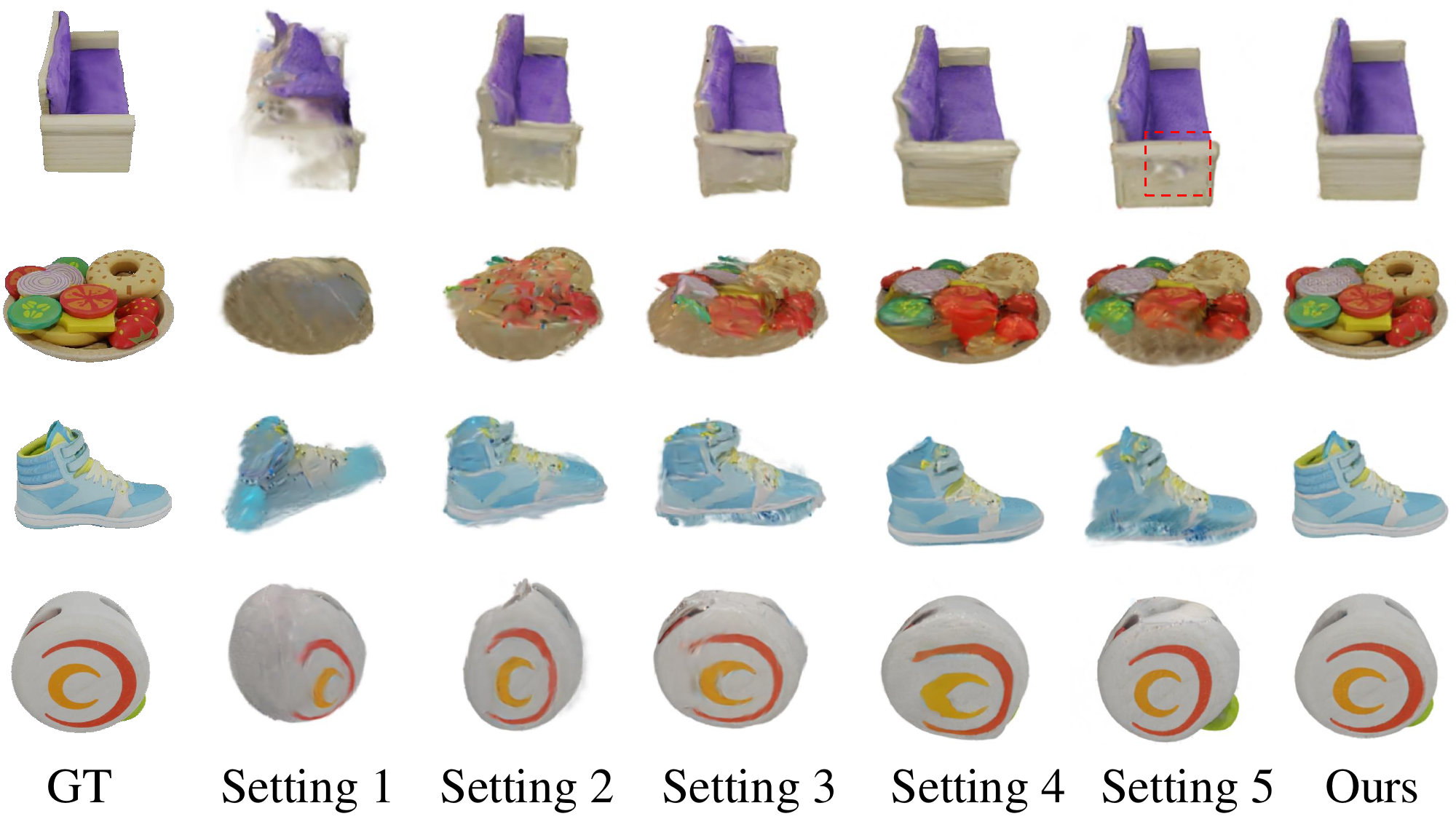}
    \caption{Qualitative results of different numbers of views. \textbf{Setting 1}: 1 clean view and 1 noisy view. \textbf{Setting 2}: 2 clean views and 1 noisy view. \textbf{Setting 3}: 3 clean views and 1 noisy view. \textbf{Setting 4}: 4 clean views and 2 noisy views. \textbf{Setting 5}: 4 clean views. }
    \label{fig:number}
    \vspace{-10pt}
\end{figure}

\textbf{Necessity of The Noisy Views}. We show qualitative comparisons of our models with and without noisy view in ~\Cref{tab:clean_gso} and ~\Cref{tab:clean_omni} on GSO and Omni3D elevation sets respectively. We can see that our model consistently achieves better quality when using noisy view images for denoising, benefiting from capturing more shape and appearance information through interacting with known clean views sufficiently. As shown in~\Cref{fig:number} setting 5, it could not generate a reasonable appearance without noisy view denoising, which is the core limation of pixel-aligned Gaussians.

\begin{table}[htbp]
\centering
\caption{Evaluation results on the GSO dataset~\cite{gso}. \checkmark means it exists, $\times$ means it doesn't exist.}
\resizebox{0.47\textwidth}{!}{
\scriptsize
\begin{tabular}{c|ccc}
\hline
 Noisy View    & PSNR $\uparrow$   & SSIM $\uparrow$ & LPIPS $\downarrow$ \\ \hline
 $\times$ & 29.985 & 0.945 & 0.070 \\ \hline
 \checkmark       & \mrkb{31.303} & \mrkb{0.946} & \mrkb{0.065} \\ \hline
\end{tabular}
}
\label{tab:clean_gso}
\end{table}
\begin{table}[htbp]
\centering
\vspace{-18pt}
\caption{Evaluation results on the Omni3D dataset~\cite{gso}.}
\vspace{-3pt}
\resizebox{0.47\textwidth}{!}{
\scriptsize
\begin{tabular}{c|ccc}
\hline
 Noisy View    & PSNR $\uparrow$   & SSIM $\uparrow$ & LPIPS $\downarrow$ \\ \hline
 $\times$ & 29.158 & 0.944 & 0.069 \\ \hline
 \checkmark       & \mrkb{31.195} & \mrkb{0.945} & \mrkb{0.067} \\ \hline
\end{tabular}
}
\label{tab:clean_omni}
\end{table}

\textbf{Number of noisy and clean views.} We present qualitative comparisons of our models with varying numbers of clean and noisy views in two different elevation sets, as detailed in~\Cref{tab:num_gso} and ~\Cref{tab:num_omni}. It reveals that the model's performance improves with an increased number of input clean images, attributable to the enhanced capture of shape and appearance information. Although novel view image denoising could promote the performance of unseen parts of the object, the computational complexity also increases significantly. So there needs to be a balance between the number of clean views and noisy views. Beyond this threshold, the presence of excessively noisy views detrimentally impacts the model's performance. As shown in the~\Cref{fig:number} setting 4, more noise view images will create more noisy Gaussian points, which will blur the image. 

\begin{table}[htbp]
\centering
\vspace{-10pt}
\caption{Evaluation results on GSO dataset. \textbf{NCV}: \textbf{N}umber of \textbf{C}lean \textbf{V}iews. \textbf{NNV}: \textbf{N}umber of \textbf{N}oisy \textbf{V}iews.}
\begin{tabular}{cc|ccc}
\hline
NCV & NNV   & PSNR $\uparrow$   & SSIM $\uparrow$ & LPIPS $\downarrow$ \\ \hline
1 & 1  & 21.668  & 0.895  & 0.167 \\
2 & 1  &  26.913 & 0.922  & 0.100 \\
3 & 1 & 29.574 & 0.938 & 0.075 \\ 
4 & 2 & \mrkc{31.256} & \mrkc{0.941} & \mrkc{0.069} \\ \hline
4 & 1 & \mrkb{31.303} & \mrkb{0.946} & \mrkb{0.065} \\ \hline
\vspace{-15pt}
\end{tabular}
\label{tab:num_gso}
\end{table}
\begin{table}[htbp]
\centering
\vspace{-15pt}
\caption{Evaluation results on Omni3D dataset.}
\begin{tabular}{cc|ccc}
\hline
NCV & NNV   & PSNR $\uparrow$   & SSIM $\uparrow$ & LPIPS $\downarrow$ \\ \hline
1 & 1  & 20.378  & 0.894  & 0.163 \\
2 & 1  &  26.108 & 0.923  & 0.097 \\
3 & 1 & 29.140 & 0.939 & 0.072 \\ 
4 & 2 & \mrkc{31.040} & \mrkc{0.941} & \mrkc{0.067} \\ \hline
4 & 1 & \mrkb{31.303} & \mrkb{0.946} & \mrkb{0.065} \\ \hline

\vspace{-20pt}
\end{tabular}
\label{tab:num_omni}
\end{table}

\begin{figure}[htbp]
    \vspace{-10pt}
    \centering
    \includegraphics[width=0.4\textwidth]{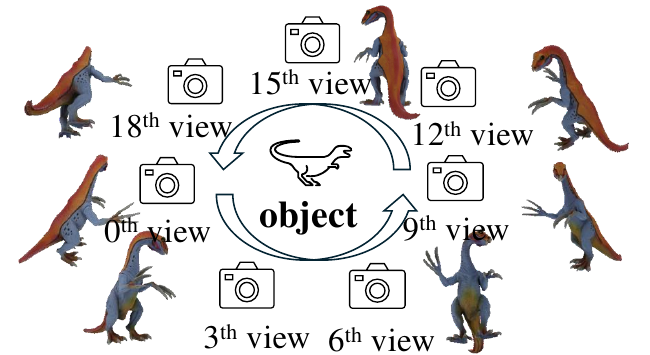} 
    \vspace{-5pt}
    \caption{Camera position demonstration.}
    \label{fig:position}
    \vspace{-5pt}
\end{figure}

\begin{figure}[htbp]
    \centering
    \includegraphics[width=1.0\columnwidth]{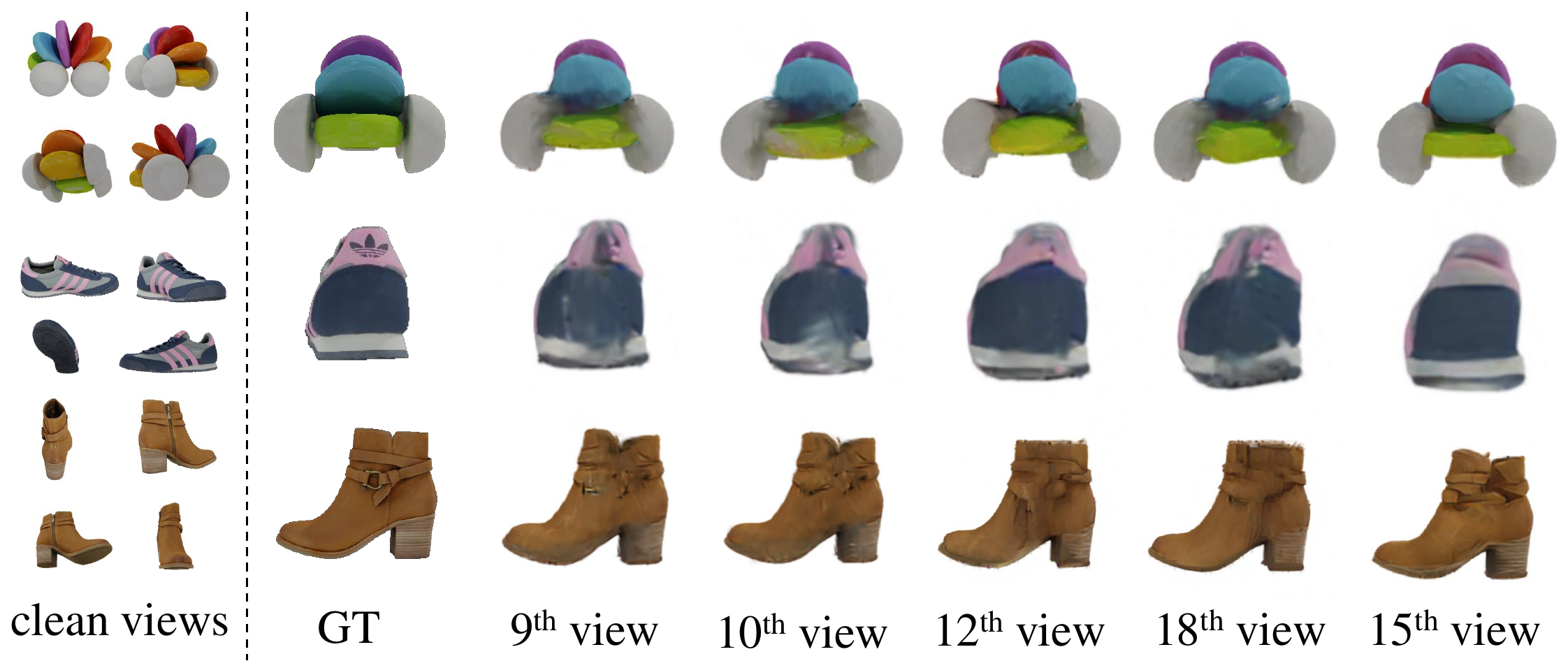}
    \caption{Visulation of the results of different positions about the noisy view. The input images are shown on the left. $i^{th}$ represent the position of the noisy view image as shown in~\Cref{fig:position}.}
    \label{fig:noisy_view}
\end{figure}

\textbf{Positional relationship between noisy view and clean views. } As shown in the ~\Cref{fig:position}, we place the object in the center and surround the cameras. We fix the clean view images and camera parameters at positions $0^{th}$, $3^{th}$, $6^{th}$, and $9^{th}$ as inputs, which cover the front of the object while not covering the back of the object. Moreover, we select positions $9^{th}$, $10^{th}$, $12^{th}$, $15^{th}$, and $18^{th}$ as the positions of the noisy view, respectively. We present quantitative comparisons of our models with varying camera poses of noisy views, as detailed in~\Cref{tab:pos_gso},~\Cref{tab:pos_omni}. When choosing $15^{th}$ as the noisy view position, the model gets the best metric. As shown in the second and third rows of~\Cref{fig:noisy_view}, choosing the $15^{th}$ view presents the best result. Even though there are some differences between this image and the ground truth, this is a reasonable phenomenon. Because the input image does not contain the parts of the object that we expect to generate. It's reasonable for the model to imagine the unseen parts and generate a detailed image. 

In conclusion, if we choose a noise view that is close to the known views, the model will take less account of parts that are not covered. As a result, it will lead to poor results in places that are not covered by the existing perspective. If we choose the positions of the noisy view and clean views that better cover the object, the model will take more account of the objects, producing better results.

\begin{table}[htbp]
\centering
\vspace{-5pt}
\caption{Evaluation results on GSO dataset. \textbf{ICV}: \textbf{I}ndex of \textbf{C}lean \textbf{V}iews. \textbf{INV}: \textbf{I}ndex of \textbf{N}oisy \textbf{V}iews}
\begin{tabular}{cc|ccc}
\hline
ICV &  INV    & PSNR $\uparrow$   & SSIM $\uparrow$ & LPIPS $\downarrow$ \\ \hline
0,3,6,9 & 9 & 31.436 & 0.947 & 0.063 \\ 
0,3,6,9 & 10 & 31.592 & 0.948 &  \mrkc{0.062} \\ 
0,3,6,9 & 12 & \mrkc{31.707} & 0.948 & 0.063 \\ 
0,3,6,9 & 18 & 31.643  & \mrkc{0.949} & \mrkc{0.062} \\ \hline
0,3,6,9 & 15 & \mrkb{32.038} & \mrkb{0.950} & \mrkb{0.061} \\ \hline
\end{tabular}
\label{tab:pos_gso}
\end{table}
\begin{table}[htbp]
\centering
\caption{Evaluation results on Omni3D dataset.}
\begin{tabular}{cc|ccc}
\hline
ICV &  INV    & PSNR $\uparrow$   & SSIM $\uparrow$ & LPIPS $\downarrow$ \\ \hline
0,3,6,9 & 9 & \mrkc{32.008} & 0.949 & 0.055 \\ 
0,3,6,9 & 10 & 31.979 & 0.948 &  \mrkc{0.055} \\ 
0,3,6,9 & 12 & 31.851 & 0.950 & 0.056 \\ 
0,3,6,9 & 18 &  31.807 & \mrkb{0.952} & 0.056 \\ \hline
0,3,6,9 & 15 & \mrkb{32.055} & \mrkc{0.951} & \mrkb{0.054} \\ \hline
\end{tabular}
\vspace{-10pt}
\label{tab:pos_omni}
\end{table}

\section{Conclusion}
In this paper, we introduce NovelGS, an innovative diffusion model designed for Gaussian Splatting (GS) using sparse-view images. Our approach employs a transformer-based network for novel view denoising, enabling the generation of 3D Gaussians. By incorporating both conditional views and noisy target views as inputs, the network predicts pixel-aligned Gaussians for each view. During the training phase, the rendered target and additional Gaussian views are supervised. In the inference phase, target views are iteratively rendered and denoised from pure noise. Our method demonstrates state-of-the-art performance in addressing the multi-view image reconstruction challenge. By generatively modeling unseen regions, NovelGS effectively reconstructs 3D objects with consistent and sharp textures. Experimental results on publicly available datasets show that NovelGS significantly outperforms existing image-to-3D frameworks, both qualitatively and quantitatively. Furthermore, we highlight the potential of NovelGS in generative tasks, such as text-to-3D and image-to-3D, by integrating it with existing multiview diffusion models.
{
    \small
    \bibliographystyle{ieeenat_fullname}
    \bibliography{main}

\begin{thebibliography}{65}
\providecommand{\natexlab}[1]{#1}
\providecommand{\url}[1]{\texttt{#1}}
\expandafter\ifx\csname urlstyle\endcsname\relax
  \providecommand{\doi}[1]{doi: #1}\else
  \providecommand{\doi}{doi: \begingroup \urlstyle{rm}\Url}\fi

\bibitem[Caron et~al.(2021)Caron, Touvron, Misra, J\'egou, Mairal, Bojanowski, and Joulin]{dino}
Mathilde Caron, Hugo Touvron, Ishan Misra, Herv\'e J\'egou, Julien Mairal, Piotr Bojanowski, and Armand Joulin.
\newblock Emerging properties in self-supervised vision transformers.
\newblock In \emph{ICCV}, 2021.

\bibitem[Chan et~al.(2021)Chan, Monteiro, Kellnhofer, Wu, and Wetzstein]{pi-gan}
Eric~R Chan, Marco Monteiro, Petr Kellnhofer, Jiajun Wu, and Gordon Wetzstein.
\newblock pi-gan: Periodic implicit generative adversarial networks for 3d-aware image synthesis.
\newblock In \emph{CVPR}, 2021.

\bibitem[Chan et~al.(2022)Chan, Lin, Chan, Nagano, Pan, De~Mello, Gallo, Guibas, Tremblay, Khamis, et~al.]{eg3d}
Eric~R Chan, Connor~Z Lin, Matthew~A Chan, Koki Nagano, Boxiao Pan, Shalini De~Mello, Orazio Gallo, Leonidas~J Guibas, Jonathan Tremblay, Sameh Khamis, et~al.
\newblock Efficient geometry-aware 3d generative adversarial networks.
\newblock In \emph{CVPR}, 2022.

\bibitem[Chang et~al.(2015)Chang, Funkhouser, Guibas, Hanrahan, Huang, Li, Savarese, Savva, Song, Su, et~al.]{shapenet}
Angel~X Chang, Thomas Funkhouser, Leonidas Guibas, Pat Hanrahan, Qixing Huang, Zimo Li, Silvio Savarese, Manolis Savva, Shuran Song, Hao Su, et~al.
\newblock Shapenet: An information-rich 3d model repository.
\newblock \emph{arXiv preprint arXiv:1512.03012}, 2015.

\bibitem[Charatan et~al.(2024)Charatan, Li, Tagliasacchi, and Sitzmann]{pixelsplat}
David Charatan, Sizhe~Lester Li, Andrea Tagliasacchi, and Vincent Sitzmann.
\newblock pixelsplat: 3d gaussian splats from image pairs for scalable generalizable 3d reconstruction.
\newblock In \emph{CVPR}, 2024.

\bibitem[Chen et~al.(2024)Chen, Xu, Esposito, Tang, and Geiger]{lara}
Anpei Chen, Haofei Xu, Stefano Esposito, Siyu Tang, and Andreas Geiger.
\newblock Lara: Efficient large-baseline radiance fields.
\newblock In \emph{ECCV}, 2024.

\bibitem[Chen et~al.(2023)Chen, Chen, Jiao, and Jia]{fantasia3d}
Rui Chen, Yongwei Chen, Ningxin Jiao, and Kui Jia.
\newblock Fantasia3d: Disentangling geometry and appearance for high-quality text-to-3d content creation.
\newblock In \emph{ICCV}, 2023.

\bibitem[Deitke et~al.(2023)Deitke, Schwenk, Salvador, Weihs, Michel, VanderBilt, Schmidt, Ehsani, Kembhavi, and Farhadi]{objaverse}
Matt Deitke, Dustin Schwenk, Jordi Salvador, Luca Weihs, Oscar Michel, Eli VanderBilt, Ludwig Schmidt, Kiana Ehsani, Aniruddha Kembhavi, and Ali Farhadi.
\newblock Objaverse: A universe of annotated 3d objects.
\newblock In \emph{CVPR}, 2023.

\bibitem[Deitke et~al.(2024)Deitke, Liu, Wallingford, Ngo, Michel, Kusupati, Fan, Laforte, Voleti, Gadre, et~al.]{objaverse-xl}
Matt Deitke, Ruoshi Liu, Matthew Wallingford, Huong Ngo, Oscar Michel, Aditya Kusupati, Alan Fan, Christian Laforte, Vikram Voleti, Samir~Yitzhak Gadre, et~al.
\newblock Objaverse-xl: A universe of 10m+ 3d objects.
\newblock \emph{NeurIPS}, 2024.

\bibitem[Downs et~al.(2022)Downs, Francis, Koenig, Kinman, Hickman, Reymann, McHugh, and Vanhoucke]{gso}
Laura Downs, Anthony Francis, Nate Koenig, Brandon Kinman, Ryan Hickman, Krista Reymann, Thomas~B McHugh, and Vincent Vanhoucke.
\newblock Google scanned objects: A high-quality dataset of 3d scanned household items.
\newblock In \emph{ICRA}, 2022.

\bibitem[Esser et~al.(2021)Esser, Rombach, and Ommer]{taming-gan}
Patrick Esser, Robin Rombach, and Bjorn Ommer.
\newblock Taming transformers for high-resolution image synthesis.
\newblock In \emph{Proceedings of the IEEE/CVF conference on computer vision and pattern recognition}, pages 12873--12883, 2021.

\bibitem[Goodfellow et~al.(2014)Goodfellow, Pouget-Abadie, Mirza, Xu, Warde-Farley, Ozair, Courville, and Bengio]{gan}
Ian Goodfellow, Jean Pouget-Abadie, Mehdi Mirza, Bing Xu, David Warde-Farley, Sherjil Ozair, Aaron Courville, and Yoshua Bengio.
\newblock Generative adversarial nets.
\newblock \emph{NeurIPS}, 2014.

\bibitem[Gupta et~al.(2023)Gupta, Xiong, Nie, Jones, and O{\u{g}}uz]{3dgen}
Anchit Gupta, Wenhan Xiong, Yixin Nie, Ian Jones, and Barlas O{\u{g}}uz.
\newblock 3dgen: Triplane latent diffusion for textured mesh generation.
\newblock \emph{arXiv preprint arXiv:2303.05371}, 2023.

\bibitem[Hong et~al.(2024)Hong, Zhang, Gu, Bi, Zhou, Liu, Liu, Sunkavalli, Bui, and Tan]{lrm}
Yicong Hong, Kai Zhang, Jiuxiang Gu, Sai Bi, Yang Zhou, Difan Liu, Feng Liu, Kalyan Sunkavalli, Trung Bui, and Hao Tan.
\newblock Lrm: Large reconstruction model for single image to 3d.
\newblock In \emph{ICLR}, 2024.

\bibitem[Jun and Nichol(2023)]{shape}
Heewoo Jun and Alex Nichol.
\newblock Shap-e: Generating conditional 3d implicit functions.
\newblock \emph{arXiv preprint arXiv:2305.02463}, 2023.

\bibitem[Kerbl et~al.(2023)Kerbl, Kopanas, Leimk{\"u}hler, and Drettakis]{gs-splatting}
Bernhard Kerbl, Georgios Kopanas, Thomas Leimk{\"u}hler, and George Drettakis.
\newblock 3d gaussian splatting for real-time radiance field rendering.
\newblock \emph{ACM Trans. Graph.}, 2023.

\bibitem[Lefaudeux et~al.(2022)Lefaudeux, Massa, Liskovich, Xiong, Caggiano, Naren, Xu, Hu, Tintore, Zhang, Labatut, Haziza, Wehrstedt, Reizenstein, and Sizov]{xFormers}
Benjamin Lefaudeux, Francisco Massa, Diana Liskovich, Wenhan Xiong, Vittorio Caggiano, Sean Naren, Min Xu, Jieru Hu, Marta Tintore, Susan Zhang, Patrick Labatut, Daniel Haziza, Luca Wehrstedt, Jeremy Reizenstein, and Grigory Sizov.
\newblock xformers: A modular and hackable transformer modelling library.
\newblock \url{https://github.com/facebookresearch/xformers}, 2022.

\bibitem[Li et~al.(2024)Li, Tan, Zhang, Xu, Luan, Xu, Hong, Sunkavalli, Shakhnarovich, and Bi]{instant3d}
Jiahao Li, Hao Tan, Kai Zhang, Zexiang Xu, Fujun Luan, Yinghao Xu, Yicong Hong, Kalyan Sunkavalli, Greg Shakhnarovich, and Sai Bi.
\newblock Instant3d: Fast text-to-3d with sparse-view generation and large reconstruction model.
\newblock In \emph{ICLR}, 2024.

\bibitem[Lin et~al.(2023)Lin, Gao, Tang, Takikawa, Zeng, Huang, Kreis, Fidler, Liu, and Lin]{magic3d}
Chen-Hsuan Lin, Jun Gao, Luming Tang, Towaki Takikawa, Xiaohui Zeng, Xun Huang, Karsten Kreis, Sanja Fidler, Ming-Yu Liu, and Tsung-Yi Lin.
\newblock Magic3d: High-resolution text-to-3d content creation.
\newblock In \emph{CVPR}, 2023.

\bibitem[Long et~al.(2024)Long, Guo, Lin, Liu, Dou, Liu, Ma, Zhang, Habermann, Theobalt, et~al.]{wonder3d}
Xiaoxiao Long, Yuan-Chen Guo, Cheng Lin, Yuan Liu, Zhiyang Dou, Lingjie Liu, Yuexin Ma, Song-Hai Zhang, Marc Habermann, Christian Theobalt, et~al.
\newblock Wonder3d: Single image to 3d using cross-domain diffusion.
\newblock \emph{CVPR}, 2024.

\bibitem[Lorensen and Cline(1998)]{marching-cube}
William~E Lorensen and Harvey~E Cline.
\newblock Marching cubes: A high resolution 3d surface construction algorithm.
\newblock In \emph{Seminal graphics: pioneering efforts that shaped the field}, pages 347--353. 1998.

\bibitem[Loshchilov(2017)]{adamw}
I Loshchilov.
\newblock Decoupled weight decay regularization.
\newblock \emph{arXiv preprint arXiv:1711.05101}, 2017.

\bibitem[Luo et~al.(2023)Luo, Rockwell, Lee, and Johnson]{cap3d}
Tiange Luo, Chris Rockwell, Honglak Lee, and Justin Johnson.
\newblock Scalable 3d captioning with pretrained models.
\newblock \emph{NeurIPS}, 2023.

\bibitem[Luo et~al.(2024)Luo, Johnson, and Lee]{cap3d_2}
Tiange Luo, Justin Johnson, and Honglak Lee.
\newblock View selection for 3d captioning via diffusion ranking.
\newblock \emph{arXiv preprint arXiv:2404.07984}, 2024.

\bibitem[Metzer et~al.(2023)Metzer, Richardson, Patashnik, Giryes, and Cohen-Or]{latent-nerf}
Gal Metzer, Elad Richardson, Or Patashnik, Raja Giryes, and Daniel Cohen-Or.
\newblock Latent-nerf for shape-guided generation of 3d shapes and textures.
\newblock In \emph{CVPR}, 2023.

\bibitem[Mildenhall et~al.(2020)Mildenhall, Srinivasan, Tancik, Barron, Ramamoorthi, and Ng]{nerf}
Ben Mildenhall, Pratul~P. Srinivasan, Matthew Tancik, Jonathan~T. Barron, Ravi Ramamoorthi, and Ren Ng.
\newblock Nerf: Representing scenes as neural radiance fields for view synthesis.
\newblock In \emph{ECCV}, 2020.

\bibitem[Nguyen-Phuoc et~al.(2019)Nguyen-Phuoc, Li, Theis, Richardt, and Yang]{hologan}
Thu Nguyen-Phuoc, Chuan Li, Lucas Theis, Christian Richardt, and Yong-Liang Yang.
\newblock Hologan: Unsupervised learning of 3d representations from natural images.
\newblock In \emph{CVPR}, 2019.

\bibitem[Nichol et~al.(2022)Nichol, Jun, Dhariwal, Mishkin, and Chen]{pointe}
Alex Nichol, Heewoo Jun, Prafulla Dhariwal, Pamela Mishkin, and Mark Chen.
\newblock Point-e: A system for generating 3d point clouds from complex prompts.
\newblock \emph{arXiv preprint arXiv:2212.08751}, 2022.

\bibitem[Niemeyer and Geiger(2021)]{giraffe}
Michael Niemeyer and Andreas Geiger.
\newblock Giraffe: Representing scenes as compositional generative neural feature fields.
\newblock In \emph{CVPR}, 2021.

\bibitem[Ntavelis et~al.(2023)Ntavelis, Siarohin, Olszewski, Wang, Gool, and Tulyakov]{auto3d}
Evangelos Ntavelis, Aliaksandr Siarohin, Kyle Olszewski, Chaoyang Wang, Luc~V Gool, and Sergey Tulyakov.
\newblock Autodecoding latent 3d diffusion models.
\newblock \emph{NeurIPS}, 2023.

\bibitem[Peebles and Xie(2023)]{dit}
William Peebles and Saining Xie.
\newblock Scalable diffusion models with transformers.
\newblock In \emph{ICCV}, 2023.

\bibitem[Poole et~al.(2022)Poole, Jain, Barron, and Mildenhall]{dreamfusion}
Ben Poole, Ajay Jain, Jonathan~T Barron, and Ben Mildenhall.
\newblock Dreamfusion: Text-to-3d using 2d diffusion.
\newblock \emph{arXiv preprint arXiv:2209.14988}, 2022.

\bibitem[Radford(2018)]{gpt-1}
Alec Radford.
\newblock Improving language understanding by generative pre-training.
\newblock 2018.

\bibitem[Rombach et~al.(2021)Rombach, Blattmann, Lorenz, Esser, and Ommer]{stable-diffusion}
Robin Rombach, Andreas Blattmann, Dominik Lorenz, Patrick Esser, and Björn Ommer.
\newblock High-resolution image synthesis with latent diffusion models, 2021.

\bibitem[Rombach et~al.(2022)Rombach, Blattmann, Lorenz, Esser, and Ommer]{sd}
Robin Rombach, Andreas Blattmann, Dominik Lorenz, Patrick Esser, and Bj{\"o}rn Ommer.
\newblock High-resolution image synthesis with latent diffusion models.
\newblock In \emph{CVPR}, 2022.

\bibitem[Shen et~al.(2023)Shen, Munkberg, Hasselgren, Yin, Wang, Chen, Gojcic, Fidler, Sharp, and Gao]{flexible-cube}
Tianchang Shen, Jacob Munkberg, Jon Hasselgren, Kangxue Yin, Zian Wang, Wenzheng Chen, Zan Gojcic, Sanja Fidler, Nicholas Sharp, and Jun Gao.
\newblock Flexible isosurface extraction for gradient-based mesh optimization.
\newblock \emph{ACM Trans. Graph.}, 2023.

\bibitem[Shi et~al.()Shi, Chen, Zhang, Liu, Xu, Wei, Chen, Zeng, and Su]{zero123plus}
Ruoxi Shi, Hansheng Chen, Zhuoyang Zhang, Minghua Liu, Chao Xu, Xinyue Wei, Linghao Chen, Chong Zeng, and Hao Su.
\newblock Zero123++: a single image to consistent multi-view diffusion base model.

\bibitem[Shue et~al.(2023)Shue, Chan, Po, Ankner, Wu, and Wetzstein]{3d-nerf}
J~Ryan Shue, Eric~Ryan Chan, Ryan Po, Zachary Ankner, Jiajun Wu, and Gordon Wetzstein.
\newblock 3d neural field generation using triplane diffusion.
\newblock In \emph{CVPR}, 2023.

\bibitem[Siddiqui et~al.(2024)Siddiqui, Alliegro, Artemov, Tommasi, Sirigatti, Rosov, Dai, and Nie{\ss}ner]{meshgpt}
Yawar Siddiqui, Antonio Alliegro, Alexey Artemov, Tatiana Tommasi, Daniele Sirigatti, Vladislav Rosov, Angela Dai, and Matthias Nie{\ss}ner.
\newblock Meshgpt: Generating triangle meshes with decoder-only transformers.
\newblock In \emph{CVPR}, 2024.

\bibitem[Singer et~al.(2022)Singer, Polyak, Hayes, Yin, An, Zhang, Hu, Yang, Ashual, Gafni, et~al.]{make-a-video}
Uriel Singer, Adam Polyak, Thomas Hayes, Xi Yin, Jie An, Songyang Zhang, Qiyuan Hu, Harry Yang, Oron Ashual, Oran Gafni, et~al.
\newblock Make-a-video: Text-to-video generation without text-video data.
\newblock \emph{arXiv preprint arXiv:2209.14792}, 2022.

\bibitem[Skorokhodov et~al.(2022)Skorokhodov, Tulyakov, Wang, and Wonka]{epigraf}
Ivan Skorokhodov, Sergey Tulyakov, Yiqun Wang, and Peter Wonka.
\newblock Epigraf: Rethinking training of 3d gans.
\newblock \emph{NeurIPS}, 2022.

\bibitem[Sohl-Dickstein et~al.(2015)Sohl-Dickstein, Weiss, Maheswaranathan, and Ganguli]{diffusion}
Jascha Sohl-Dickstein, Eric Weiss, Niru Maheswaranathan, and Surya Ganguli.
\newblock Deep unsupervised learning using nonequilibrium thermodynamics.
\newblock In \emph{ICML}. PMLR, 2015.

\bibitem[Sun et~al.(2024)Sun, Zhang, Shao, Wang, Liu, Xie, and Liu]{dreamcraft3d}
Jingxiang Sun, Bo Zhang, Ruizhi Shao, Lizhen Wang, Wen Liu, Zhenda Xie, and Yebin Liu.
\newblock Dreamcraft3d: Hierarchical 3d generation with bootstrapped diffusion prior.
\newblock In \emph{ICLR}, 2024.

\bibitem[Szymanowicz et~al.(2023)Szymanowicz, Rupprecht, and Vedaldi]{viewset}
Stanislaw Szymanowicz, Christian Rupprecht, and Andrea Vedaldi.
\newblock Viewset diffusion:(0-) image-conditioned 3d generative models from 2d data.
\newblock In \emph{ICCV}, 2023.

\bibitem[Szymanowicz et~al.(2024)Szymanowicz, Rupprecht, and Vedaldi]{splatter}
Stanislaw Szymanowicz, Christian Rupprecht, and Andrea Vedaldi.
\newblock Splatter image: Ultra-fast single-view 3d reconstruction.
\newblock In \emph{CVPR}, 2024.

\bibitem[Tang et~al.(2024{\natexlab{a}})Tang, Chen, Chen, Wang, Zeng, and Liu]{lgm}
Jiaxiang Tang, Zhaoxi Chen, Xiaokang Chen, Tengfei Wang, Gang Zeng, and Ziwei Liu.
\newblock Lgm: Large multi-view gaussian model for high-resolution 3d content creation.
\newblock \emph{ECCV}, 2024{\natexlab{a}}.

\bibitem[Tang et~al.(2024{\natexlab{b}})Tang, Ren, Zhou, Liu, and Zeng]{dreamgaussian}
Jiaxiang Tang, Jiawei Ren, Hang Zhou, Ziwei Liu, and Gang Zeng.
\newblock Dreamgaussian: Generative gaussian splatting for efficient 3d content creation.
\newblock In \emph{ICLR}, 2024{\natexlab{b}}.

\bibitem[Tochilkin et~al.(2024)Tochilkin, Pankratz, Liu, Huang, , Letts, Li, Liang, Laforte, Jampani, and Cao]{triposr}
Dmitry Tochilkin, David Pankratz, Zexiang Liu, Zixuan Huang, , Adam Letts, Yangguang Li, Ding Liang, Christian Laforte, Varun Jampani, and Yan-Pei Cao.
\newblock Triposr: Fast 3d object reconstruction from a single image.
\newblock \emph{arXiv preprint arXiv:2403.02151}, 2024.

\bibitem[Vaswani(2017)]{transformer}
A Vaswani.
\newblock Attention is all you need.
\newblock \emph{NeurIPS}, 2017.

\bibitem[Wang et~al.(2024{\natexlab{a}})Wang, Tan, Bi, Xu, Luan, Sunkavalli, Wang, Xu, and Zhang]{pf-lrm}
Peng Wang, Hao Tan, Sai Bi, Yinghao Xu, Fujun Luan, Kalyan Sunkavalli, Wenping Wang, Zexiang Xu, and Kai Zhang.
\newblock Pf-lrm: Pose-free large reconstruction model for joint pose and shape prediction.
\newblock In \emph{ICLR}, 2024{\natexlab{a}}.

\bibitem[Wang et~al.(2004)Wang, Bovik, Sheikh, and Simoncelli]{ssim}
Zhou Wang, Alan~C Bovik, Hamid~R Sheikh, and Eero~P Simoncelli.
\newblock Image quality assessment: from error visibility to structural similarity.
\newblock \emph{IEEE transactions on image processing}, 2004.

\bibitem[Wang et~al.(2023)Wang, Lu, Wang, Bao, Li, Su, and Zhu]{prolificdreamer}
Zhengyi Wang, Cheng Lu, Yikai Wang, Fan Bao, Chongxuan Li, Hang Su, and Jun Zhu.
\newblock Prolificdreamer: High-fidelity and diverse text-to-3d generation with variational score distillation.
\newblock In \emph{NeurIPS}, 2023.

\bibitem[Wang et~al.(2024{\natexlab{b}})Wang, Wang, Chen, Xiang, Chen, Yu, Li, Su, and Zhu]{CRM}
Zhengyi Wang, Yikai Wang, Yifei Chen, Chendong Xiang, Shuo Chen, Dajiang Yu, Chongxuan Li, Hang Su, and Jun Zhu.
\newblock Crm: Single image to 3d textured mesh with convolutional reconstruction model.
\newblock In \emph{ECCV}, 2024{\natexlab{b}}.

\bibitem[Wei et~al.(2024)Wei, Zhang, Bi, Tan, Luan, Deschaintre, Sunkavalli, Su, and Xu]{meshlrm}
Xinyue Wei, Kai Zhang, Sai Bi, Hao Tan, Fujun Luan, Valentin Deschaintre, Kalyan Sunkavalli, Hao Su, and Zexiang Xu.
\newblock Meshlrm: Large reconstruction model for high-quality mesh.
\newblock \emph{arXiv preprint arXiv:2404.12385}, 2024.

\bibitem[Wu et~al.(2023)Wu, Zhang, Fu, Wang, Ren, Pan, Wu, Yang, Wang, Qian, et~al.]{omni3d}
Tong Wu, Jiarui Zhang, Xiao Fu, Yuxin Wang, Jiawei Ren, Liang Pan, Wayne Wu, Lei Yang, Jiaqi Wang, Chen Qian, et~al.
\newblock Omniobject3d: Large-vocabulary 3d object dataset for realistic perception, reconstruction and generation.
\newblock In \emph{CVPR}, 2023.

\bibitem[Xu et~al.(2023)Xu, Wang, Cheng, Cao, Shan, Qie, and Gao]{dream3d}
Jiale Xu, Xintao Wang, Weihao Cheng, Yan-Pei Cao, Ying Shan, Xiaohu Qie, and Shenghua Gao.
\newblock Dream3d: Zero-shot text-to-3d synthesis using 3d shape prior and text-to-image diffusion models.
\newblock In \emph{CVPR}, 2023.

\bibitem[Xu et~al.(2024{\natexlab{a}})Xu, Cheng, Gao, Wang, Gao, and Shan]{instantmesh}
Jiale Xu, Weihao Cheng, Yiming Gao, Xintao Wang, Shenghua Gao, and Ying Shan.
\newblock Instantmesh: Efficient 3d mesh generation from a single image with sparse-view large reconstruction models.
\newblock \emph{arXiv preprint arXiv:2404.07191}, 2024{\natexlab{a}}.

\bibitem[Xu et~al.(2022)Xu, Peng, Yang, Shen, and Zhou]{3d-aware-gan}
Yinghao Xu, Sida Peng, Ceyuan Yang, Yujun Shen, and Bolei Zhou.
\newblock 3d-aware image synthesis via learning structural and textural representations.
\newblock In \emph{CVPR}, pages 18430--18439, 2022.

\bibitem[Xu et~al.(2024{\natexlab{b}})Xu, Shi, Yifan, Chen, Yang, Peng, Shen, and Wetzstein]{grm}
Yinghao Xu, Zifan Shi, Wang Yifan, Hansheng Chen, Ceyuan Yang, Sida Peng, Yujun Shen, and Gordon Wetzstein.
\newblock Grm: Large gaussian reconstruction model for efficient 3d reconstruction and generation.
\newblock \emph{arXiv preprint arXiv:2403.14621}, 2024{\natexlab{b}}.

\bibitem[Xu et~al.(2024{\natexlab{c}})Xu, Tan, Luan, Bi, Wang, Li, Shi, Sunkavalli, Wetzstein, Xu, et~al.]{dmv3d}
Yinghao Xu, Hao Tan, Fujun Luan, Sai Bi, Peng Wang, Jiahao Li, Zifan Shi, Kalyan Sunkavalli, Gordon Wetzstein, Zexiang Xu, et~al.
\newblock Dmv3d: Denoising multi-view diffusion using 3d large reconstruction model.
\newblock \emph{ICLR}, 2024{\natexlab{c}}.

\bibitem[Zhang et~al.(2024{\natexlab{a}})Zhang, Cheng, Yang, Wang, Zhao, Tang, Chen, and Guo]{gaussiancube}
Bowen Zhang, Yiji Cheng, Jiaolong Yang, Chunyu Wang, Feng Zhao, Yansong Tang, Dong Chen, and Baining Guo.
\newblock Gaussiancube: Structuring gaussian splatting using optimal transport for 3d generative modeling.
\newblock \emph{arXiv preprint arXiv:2403.19655}, 2024{\natexlab{a}}.

\bibitem[Zhang et~al.(2024{\natexlab{b}})Zhang, Song, Wei, Chen, Lu, and Tang]{geolrm}
Chubin Zhang, Hongliang Song, Yi Wei, Yu Chen, Jiwen Lu, and Yansong Tang.
\newblock Geolrm: Geometry-aware large reconstruction model for high-quality 3d gaussian generation.
\newblock \emph{arXiv preprint arXiv:2406.15333}, 2024{\natexlab{b}}.

\bibitem[Zhang et~al.(2022)Zhang, Kolkin, Bi, Luan, Xu, Shechtman, and Snavely]{deferred}
Kai Zhang, Nick Kolkin, Sai Bi, Fujun Luan, Zexiang Xu, Eli Shechtman, and Noah Snavely.
\newblock Arf: Artistic radiance fields.
\newblock In \emph{ECCV}, 2022.

\bibitem[Zhang et~al.(2024{\natexlab{c}})Zhang, Bi, Tan, Xiangli, Zhao, Sunkavalli, and Xu]{gs-lrm}
Kai Zhang, Sai Bi, Hao Tan, Yuanbo Xiangli, Nanxuan Zhao, Kalyan Sunkavalli, and Zexiang Xu.
\newblock Gs-lrm: Large reconstruction model for 3d gaussian splatting.
\newblock \emph{arXiv preprint arXiv:2404.19702}, 2024{\natexlab{c}}.

\bibitem[Zhang et~al.(2018)Zhang, Isola, Efros, Shechtman, and Wang]{lpips}
Richard Zhang, Phillip Isola, Alexei~A Efros, Eli Shechtman, and Oliver Wang.
\newblock The unreasonable effectiveness of deep features as a perceptual metric.
\newblock In \emph{CVPR}, 2018.

\end{thebibliography}
}
\maketitlesupplementary

In this supplementary material, we provide additional details and experiments not included in the main paper due to space limitations.
\begin{itemize}
    \item \Cref{supp:dataset}: Details of the dataset filter.
    \item \Cref{supp:model}: Details of our NovelGS models.
    \item \Cref{supp:gaussian}: Details of 3D Gaussian Parameterization.
    \item \Cref{supp:quantitative}: Quantitative evaluation on the Omni3D~\cite{omni3d} benchmark views. 
    \item \Cref{supp:exploration}: Exploration for more input views. 
    \item \Cref{supp:visual}: Additional visual results of NovelGS.
\end{itemize}

\section{Data Preparation}\label{supp:dataset}
The training dataset utilized in this study comprises multi-view images generated from the Objaverse~\cite{objaverse} dataset. We render images at a resolution of 512 × 512 pixels for each object, along with corresponding depth and normal maps, from 32 randomly selected viewpoints. We employ a filtered subset of high-quality images to enhance the quality of our model training. The filtering process aims to exclude objects that meet any of the following criteria:

\begin{enumerate}
\item Absence of texture maps.
\item Rendered images that occupy less than 10\% of the view from any perspective.
\item Inclusion of multiple distinct objects.
\item Lack of caption information as provided by the Cap3D dataset~\cite{cap3d,cap3d_2}.
\item Objects classified as low quality. The designation of ``low quality" is based on the presence of specific tags, such as ``lowpoly" and its variants (e.g., "low\_poly"), within the metadata.
\end{enumerate}

By applying these filtering criteria, we successfully curated approximately 270k high-quality instances from an initial pool of 800k objects in the Objaverse dataset.

\section{Additional Model Details}\label{supp:model}
Inspired by~\cite{gs-lrm, meshlrm} which claim the two-stage training scheme cuts the computing cost significantly, we train our model in two stages, as explained in our training setting part. Specifically, we initially pretrain our model using images with a resolution of 256 × 256 pixels until convergence is achieved. Subsequently, we perform fine-tuning for a reduced number of iterations utilizing images at a resolution of 512 × 512 pixels. The fine-tuning phase employs the same model architecture and initializes the model using the weights obtained from pre-training; however, it processes a greater number of tokens compared to the pre-training phase. Each stage of the model parameter dimension details are shown in the~\Cref{tab:parameter}. 
\begin{table*}[htbp]
\centering
\caption{Details of NovelGS model parameters in different training stages.}
\begin{tabular}{llcc}
\hline
\multicolumn{2}{l}{Parameter}                                                  & \multicolumn{1}{c}{Stage 1} & \multicolumn{1}{c}{Stage 2} \\ \hline
\multicolumn{1}{l|}{\multirow{2}{*}{Input}}               & Resolution          & 256                          & 512                         \\ \cline{2-4} 
\multicolumn{1}{l|}{}                                     & Views (clean/noisy) & 4/1                          & 4/1                         \\ \hline
\multicolumn{1}{l|}{\multirow{2}{*}{Time Embedding}}      & Frequency dim       & 256                          & 256                         \\ \cline{2-4} 
\multicolumn{1}{l|}{}                                     & Embedding dim       & 768                          & 768                         \\ \hline
\multicolumn{1}{l|}{Convolution}                          & Stride              & 8                            & 8                           \\ \hline
\multicolumn{1}{l|}{}                                     & Kernel size         & 8                            & 8                           \\ \hline
\multicolumn{1}{l|}{\multirow{2}{*}{Denoiser}}            & Transformer dim     & 768                          & 768                         \\ \cline{2-4} 
\multicolumn{1}{l|}{}                                     & Transformer layers  & 24                           & 24                          \\ \hline
\multicolumn{1}{l|}{\multirow{2}{*}{Gaussian Render}}     & Render size         & 256                          & 512                         \\ \cline{2-4} 
\multicolumn{1}{l|}{}                                     & Patch size          & 256                          & 512                         \\ \hline
\multicolumn{1}{l|}{\multirow{4}{*}{Diffusion Scheduler}} & Num timesteps       & 1000                         & 1000                        \\ \cline{2-4} 
\multicolumn{1}{l|}{}                                     & Beta schedule       & squaredcos\_cap\_v2          & squaredcos\_cap\_v2         \\ \cline{2-4} 
\multicolumn{1}{l|}{}                                     & Beta start          & 0.0001                       & 0.0001                      \\ \cline{2-4} 
\multicolumn{1}{l|}{}                                     & Beta end            & 0.02                         & 0.02                        \\ \hline
\end{tabular}\label{tab:parameter}
\end{table*}

\section{3D Gaussian Parameterization}\label{supp:gaussian}
As detailed explained in the previous work~\cite{gs-lrm, grm, lgm}, 3D Gaussians represent an unstructured explicit form; in contrast to the structural implicit representation used in Triplane-NeRF, the way output parameters are parameterized can significantly influence the model's convergence. This difference in representation means that the choice of parameterization can impact how effectively the model learns and optimizes during training, potentially affecting both the speed of convergence and the quality of the final output. We provide a detailed discussion on how we implement the parameterization to ensure reproducibility.

\textbf{Rotation.} To ensure that these quaternions represent valid rotations, we apply L2 normalization as an activation function. This process transforms the unnormalized quaternions into unit quaternions, which are essential for representing rotations in 3D space accurately. 

\textbf{RGB.} We utilize the sigmoid activation function to interpret the model's output as the zero-order Spherical Harmonics coefficients, which are utilized in the Gaussian Splatting implementation~\cite{gs-splatting}.  By focusing on zero-order coefficients, we streamline the representation while still achieving satisfactory results for our current objectives. 

\textbf{XYZ.} We don't predict the XYZ directly. Our model predicts the depth $G_{depth}$ and projects the depth value to XYZ. We use the sigmoid activation defined as $\Phi(x)=1/(1+\exp(-x))$.The Gaussian center is parameterized as 
\begin{gather}
    xyz=ray_o + t*(ray_d) \\
    w = \Phi (G_{depth}) \\
    t = (1 - w) * d_{near} + w * d_{far}
\end{gather}
We utilize $z_{near}=0.1$ and $z_{far}=4.5$ to clip the predicted XYZ to the unit space ${[-1,1]}^3$.

\textbf{Opacity.} We utilize the sigmoid function to map the range to $R^+$ and (0, 1).

\textbf{Scale.} For the scale of the Gaussian Splatting, we utilize the sigmoid function as the activation function. In addition to applying activations, we aimed to ensure that the initial output of our model is close to a fixed value. To achieve this, we introduced a constant bias to the output of the transformer, effectively shifting the initialization of the model. This adjustment helps guide the model toward a more favorable starting point for training. Specially, 
\begin{gather}
    scale = s_{min} + (s_{max} - s_{min}) * \Phi(G_{scale})
\end{gather}
where $s_{min}=0.005$ and $s_{max}=0.02$.

\section{Additional Quantitative Results}\label{supp:quantitative}
In this part, we report the performance of our model in a new setting of Omni3D~\cite{omni3d} as shown in~\Cref{table:recon_omni_rand}. Given that Omni3D incorporates benchmark views randomly sampled from the upper semi-sphere of each object, we randomly select 16 views to construct an additional image evaluation set specifically for Omni3D following InstantMesh~\cite{instantmesh}. The experimental results demonstrate the superiority of our model's results compared to the previous models.

\begin{table}[htbp]
\centering
\vspace{-5pt}
\caption{Evaluation results on the Omni3D benchmark views.}
\vspace{-5pt}
\begin{tabular}{l|ccc}
\hline
& \multicolumn{3}{c}{OmniObject3D~\cite{omni3d}} \\
& PSNR $\uparrow$   & SSIM $\uparrow$ & LPIPS $\downarrow$ \\ \hline
LGM ~\cite{lgm}  & 20.146     & 0.866    & 0.149    \\
InstantMesh~\cite{instantmesh}   & 18.864     & 0.888    & \mrkc{0.136}    \\
GeoLRM~\cite{geolrm} & 19.783     & 0.887   &  \mrkb{0.133}   \\ 
LaRa~\cite{lara}    &  21.857    & \mrkb{0.898}    &  0.143  \\ \hline
Ours    &  \mrkb{23.097}    &  \mrkc{0.889}  & \mrkc{0.136}  \\ \hline
\end{tabular}
\vspace{-10pt}
\label{table:recon_omni_rand}
\end{table}

\section{Exploration for More Input Views}\label{supp:exploration}
In this part, we explore the model's performance and GPU inference memory with more input views. We could find the performance of the model improves as the number of input views increases; however, this also leads to an increase in the GPU memory requirements of the model. This trade-off highlights the need to balance performance gains with the available hardware resources. (In the main experiments in the paper, we adopt 4 clean view images and 1 noisy image. When selecting the batch size as 1, the GPU memory is about 33.852 GB at the training stage. Considering the hardware resources, we keep this setting as default. )
\begin{table}[htbp]
\centering
\vspace{-5pt}
\caption{Evaluation results on GSO dataset. \textbf{NCV}: \textbf{N}umber of \textbf{C}lean \textbf{V}iews. \textbf{NNV}: \textbf{N}umber of \textbf{N}oisy \textbf{V}iews. Memory (GB): GPU Memory Usage in Inference Stage. }
\vspace{-5pt}
\scalebox{0.85}{
\begin{tabular}{cc|cccc}
\hline
NCV & NNV   & PSNR $\uparrow$   & SSIM $\uparrow$ & LPIPS $\downarrow$ & Memory (GB)\\ \hline
1 & 1  & 21.668  & 0.895  & 0.167& 4.131\\
2 & 1  &  26.913 & 0.922  & 0.100 & 5.021\\
3 & 1 & 29.574 & 0.938 & 0.075 & 6.049\\ 
4 & 2 & 31.256 & 0.941 & 0.069 & 7.880\\ 
4 & 1 & 31.303 & 0.946 & 0.065 & 6.874\\ 
5 & 1 & 32.440 & 0.952 & 0.057& 7.879\\ 
6 & 1 & 33.177 & 0.955 & 0.053& 8.911\\ 
7 & 1 & \mrkc{33.716} & \mrkc{0.957} & \mrkc{0.051} & 9.916\\ 
8 & 1 & \mrkb{34.259} & \mrkb{0.960} & \mrkb{0.048} & 10.932\\ \hline
\end{tabular}}
\vspace{-10pt}
\label{tab:num_gso_more}
\end{table}
\begin{table}[htbp]
\centering
\vspace{-10pt}
\caption{Evaluation results on Omni3D dataset.}
\vspace{-5pt}
\scalebox{0.85}{
\begin{tabular}{cc|cccc}
\hline
NCV & NNV   & PSNR $\uparrow$   & SSIM $\uparrow$ & LPIPS $\downarrow$ & Memory (GB)\\ \hline
1 & 1  & 20.378  & 0.894  & 0.163 &4.131\\
2 & 1  &  26.108 & 0.923  & 0.097&5.021\\
3 & 1 & 29.140 & 0.939 & 0.072&6.049 \\ 
4 & 2 & 31.040 & 0.941 & 0.067 &7.880\\ 
4 & 1 & 31.303 & 0.946 & 0.065 & 6.874\\ 
5 & 1 & 32.396 & 0.953 & 0.054&7.879\\ 
6 & 1 & 33.114 & 0.956 & 0.051 &8.911\\ 
7 & 1 & \mrkc{33.665} & \mrkc{0.957} & \mrkc{0.048} & 9.916\\ 
8 & 1 & \mrkb{34.134} & \mrkb{0.960} & \mrkb{0.046} &10.932\\ \hline
\end{tabular}}
\label{tab:num_omni_more}
\vspace{-5pt}
\end{table}

\section{Additional Visual Results}\label{supp:visual}

\begin{figure*}[htbp]
    \centering
    \includegraphics[width=1.0\linewidth]{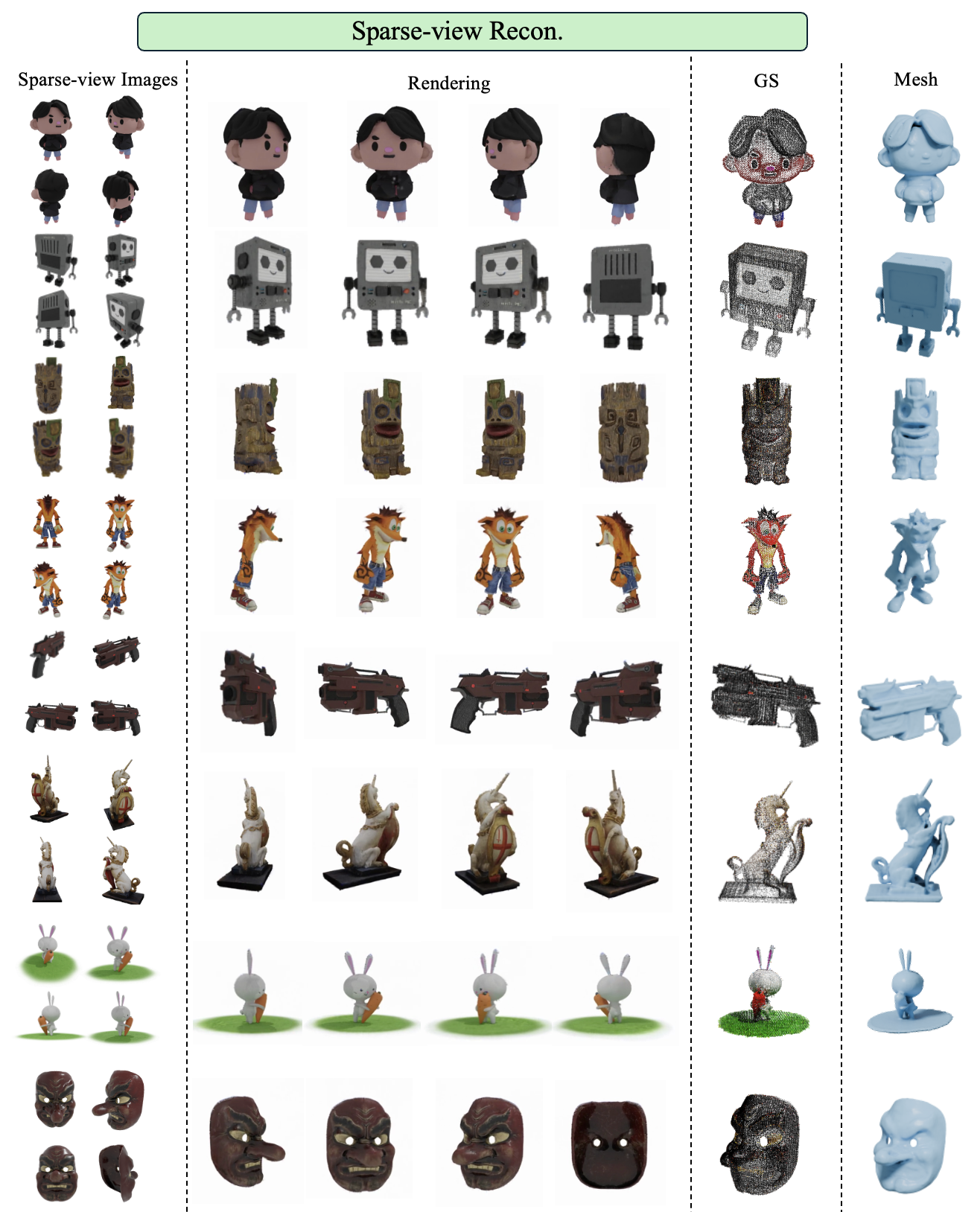}
    \caption{\textbf{High-fidelity 3D assets} produced by \textbf{NovelGS} through sparse-view images.}
    \label{fig:sparse-to-3d}
\end{figure*}

\begin{figure*}[htbp]
    \centering
    \includegraphics[width=1.0\linewidth]{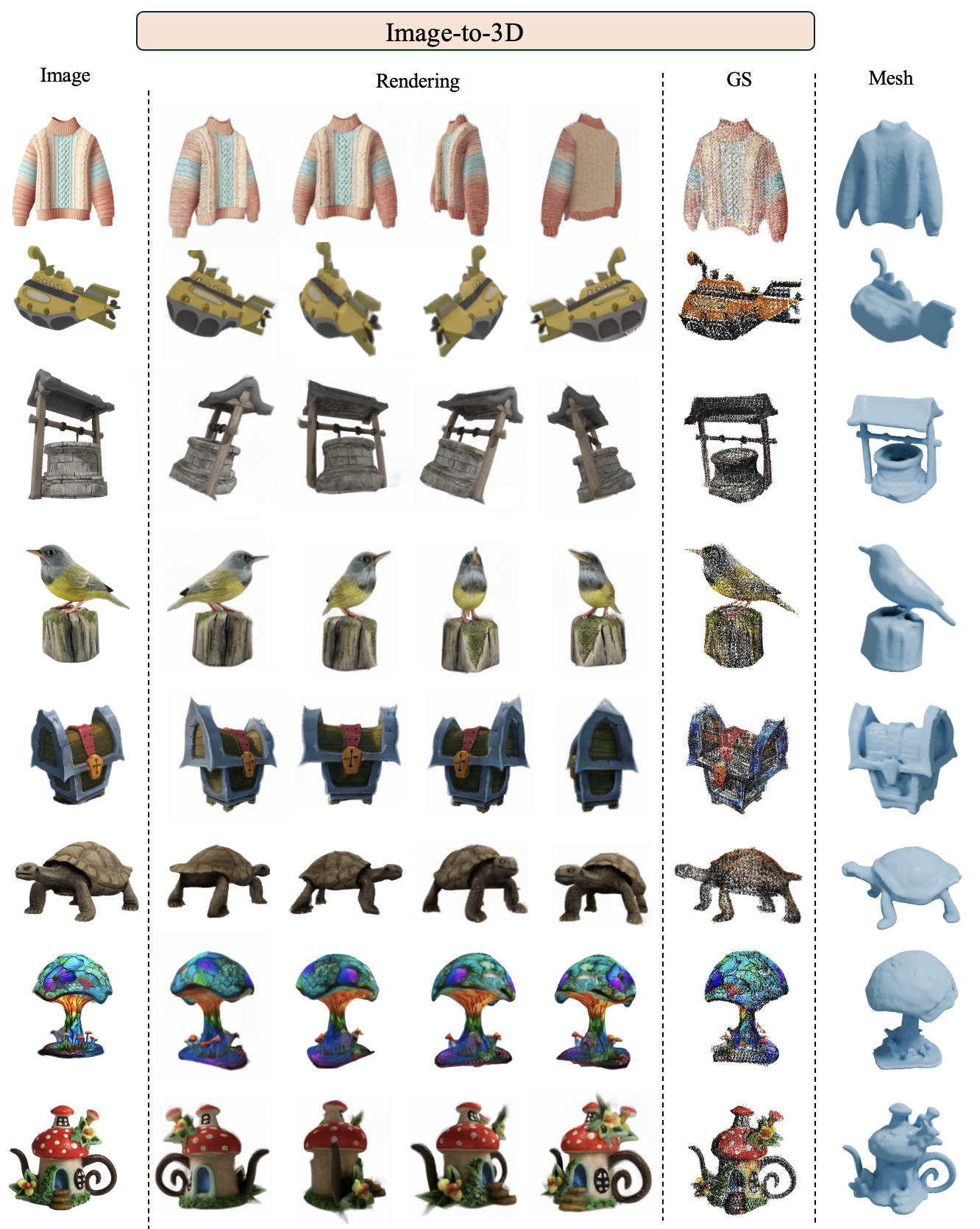}
    \caption{\textbf{High-fidelity 3D assets} produced by \textbf{NovelGS} through a single image.}
    \label{fig:img-to-3d}
\end{figure*}

\begin{figure*}[htbp]
    \centering
    \includegraphics[width=1.0\linewidth]{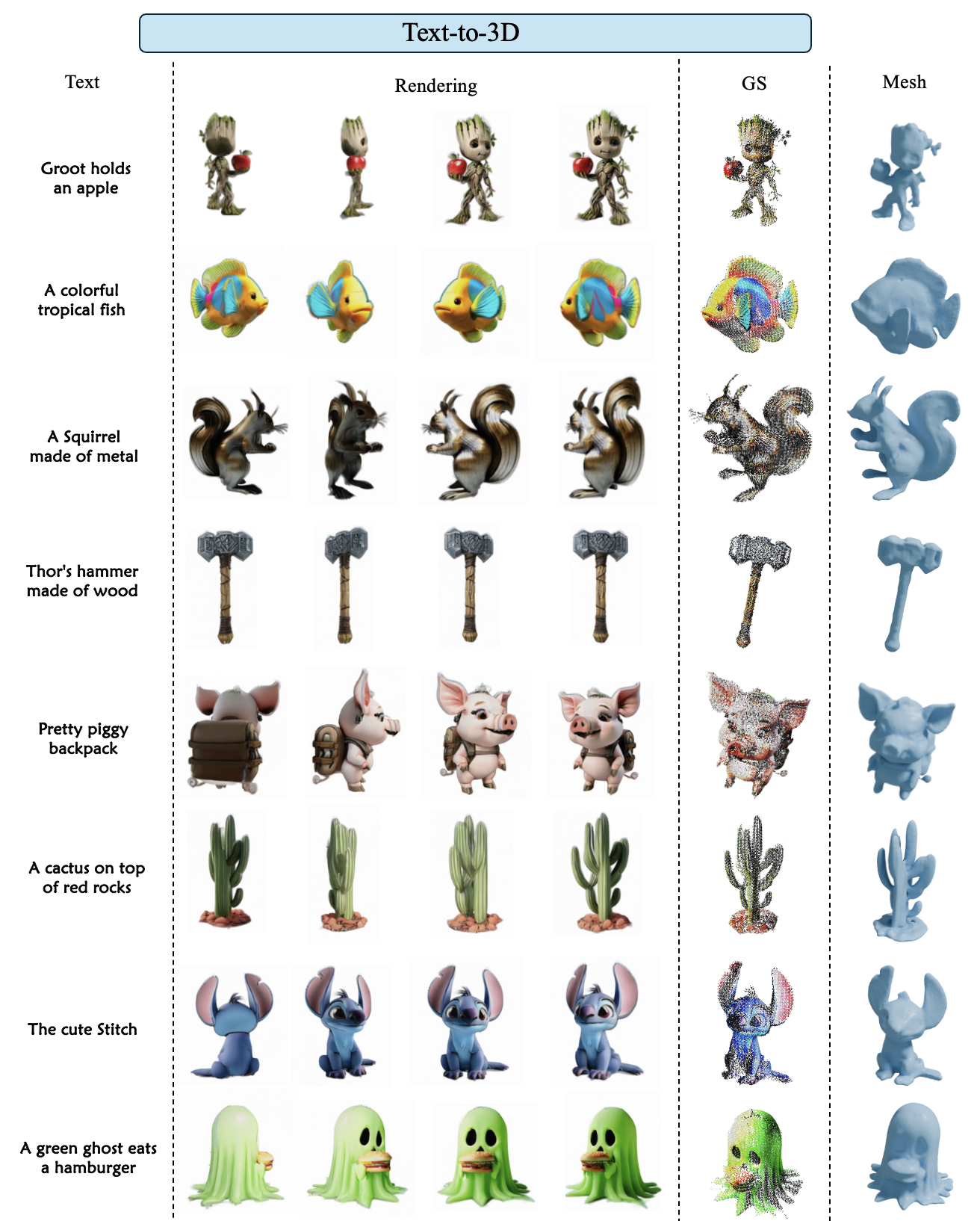}
    \caption{\textbf{High-fidelity 3D assets} produced by \textbf{NovelGS} through a sentence.}
    \label{fig:text-to-3d}
\end{figure*}

As shown in~\Cref{fig:sparse-to-3d},~\Cref{fig:img-to-3d} and~\Cref{fig:text-to-3d}, we show more high-fidelity 3D assets generated by our NovelGS.

\clearpage


\end{document}